\documentclass[journal,twoside]{IEEEtran}

\ifCLASSINFOpdf
   \usepackage[pdftex]{graphicx}
\else
\fi

\usepackage{url}
\usepackage[binary-units=true]{siunitx}
\usepackage{xcolor}
\usepackage{tikz}
\usepackage[mode=buildnew]{standalone}
\usepackage{siunitx}
\usepackage{amssymb}
\usepackage{amsmath}
\usepackage{booktabs}
\usepackage{tabularx}
\usepackage{multirow}
\usepackage{enumitem}  
\usepackage{hyperref}
\hypersetup{
    colorlinks=true,
    linkcolor=black,
    filecolor=magenta,      
    urlcolor=black,
}

\newcommand*\circled[1]{\tikz[baseline=(char.base)]{
            \node[shape=circle,fill,inner sep=1pt] (char) {\textcolor{white}{#1}};}}

\begin{document}

\title{Fully Onboard AI-powered Human-Drone Pose Estimation on Ultra-low Power Autonomous Flying Nano-UAVs}

\author{Daniele Palossi,
        Nicky Zimmerman,
        Alessio Burrello,
        Francesco Conti,
        Hanna M\"{u}ller,
        Luca Maria Gambardella,
        Luca Benini,
        Alessandro Giusti,
        J\'er\^ome Guzzi%
\thanks{This work has been partially funded by the Swiss National Science Foundation (SNSF) Spark (grant no. 190880), by the Swiss National Centre of Competence in Research (NCCR) Robotics, and by the EU H2020 project ALOHA (grant no. 780788).}%
\thanks{D. Palossi, N. Zimmerman, L. M. Gambardella, A. Giusti, and J. Guzzi are with the Dalle Molle Institute for Artificial Intelligence (IDSIA), University of Lugano \& SUPSI, Via La Santa 1, 6900 Lugano, Switzerland (e-mail: name.surname@idsia.ch).}%
\thanks{A. Burrello, F. Conti, and L. Benini are with the Department of Electrical, Electronic and Information Engineering (DEI) of University of Bologna, Viale del Risorgimento 2, 40136 Bologna, Italy (e-mail: name.surname@unibo.it).}%
\thanks{D. Palossi, H. M\"uller, and L. Benini are with the Integrated Systems Laboratory (IIS) of ETH Z\"urich, ETZ, Gloriastrasse 35, 8092 Z\"urich, Switzerland (e-mail: name.surname@iis.ee.ethz.ch).}%
\thanks{This work has been submitted to the IEEE for possible publication. Copyright may be transferred without notice, after which this version may no longer be accessible.}}

% The paper headers
\markboth{}{{D. Palossi \MakeLowercase{\textit{et al.}}: AI-powered Human-Drone Pose Estimation Aboard Ultra-low Power Autonomous Flying Nano-UAVs.}}

% make the title area
\maketitle

\begin{abstract}
Artificial intelligence-powered pocket-sized air robots have the potential to revolutionize the Internet-of-Things ecosystem, acting as autonomous, unobtrusive, and ubiquitous smart sensors.
With a few \SI{}{\cm\squared} form-factor, nano-sized unmanned aerial vehicles (UAVs) are the natural befit for indoor human-drone interaction missions, as the pose estimation task we address in this work.
However, this scenario is challenged by the nano-UAVs' limited payload and computational power that severely relegates the onboard brain to the sub-100 \SI{}{\milli\watt} microcontroller unit-class.
Our work stands at the intersection of the novel parallel ultra-low-power (PULP) architectural paradigm and our general development methodology for deep neural network (DNN) visual pipelines, i.e., covering from perception to control.
Addressing the DNN model design, from training and dataset augmentation to 8-bit quantization and deployment, we demonstrate how a PULP-based processor, aboard a nano-UAV, is sufficient for the real-time execution (up to \SI{135}{frame/\second}) of our novel DNN, called PULP-Frontnet.
We showcase how, scaling our model's memory and computational requirement, we can significantly improve the onboard inference (top energy efficiency of \SI{0.43}{\milli\joule/frame}) with no compromise in the quality-of-result vs. a resource-unconstrained baseline (i.e., full-precision DNN).
Field experiments demonstrate a closed-loop top-notch autonomous navigation capability, with a heavily resource-constrained 27-grams Crazyflie 2.1 nano-quadrotor.
Compared against the control performance achieved using an ideal sensing setup, onboard relative pose inference yields excellent drone behavior in terms of median absolute errors, such as positional (onboard: \SI{41}{\cm}, ideal: \SI{26}{\cm}) and angular (onboard: \ang{3.7}, ideal: \ang{4.1}).
\end{abstract}

\begin{IEEEkeywords}
Autonomous UAV, Convolutional Neural Networks, Ultra-low-power, Nano-UAV, Artificial Intelligence.\end{IEEEkeywords}

\IEEEpeerreviewmaketitle

\section{Introduction}\label{sec:introduction}

\IEEEPARstart{A}{utonomous} pocket-sized unmanned aerial vehicles (UAVs) powered by artificial intelligence (AI) represent a game-changer element in the Internet of Things (IoT) domain, with applications ranging from search and rescue missions to human-drone interaction (HDI) and precision agriculture~\cite{survey2019UAVapplications,iotUAV_survey}.
Nano-sized UAVs, with a sub-ten centimeters form-factor and a few tens of grams in weight, have the potential to enable exciting use cases, out of reach for bulkier aircraft~\cite{future2015}.
Enabling powerful perception algorithms on these compact and versatile platforms would allow for their adoption as unobtrusive and mobile ``smart sensors''. 
Therefore, maximizing their sense-and-act capabilities, such as autonomously flying where their presence is more valuable, such as near humans~\cite{mantegazza2019vision}.
Additionally, these small-scale UAVs can also perform onboard data analytics to preselect vital information transmitted to the IoT backbone~\cite{avasalcai2020edge}.

\begin{figure}[t]
    \centering
    \includegraphics[width=\linewidth]{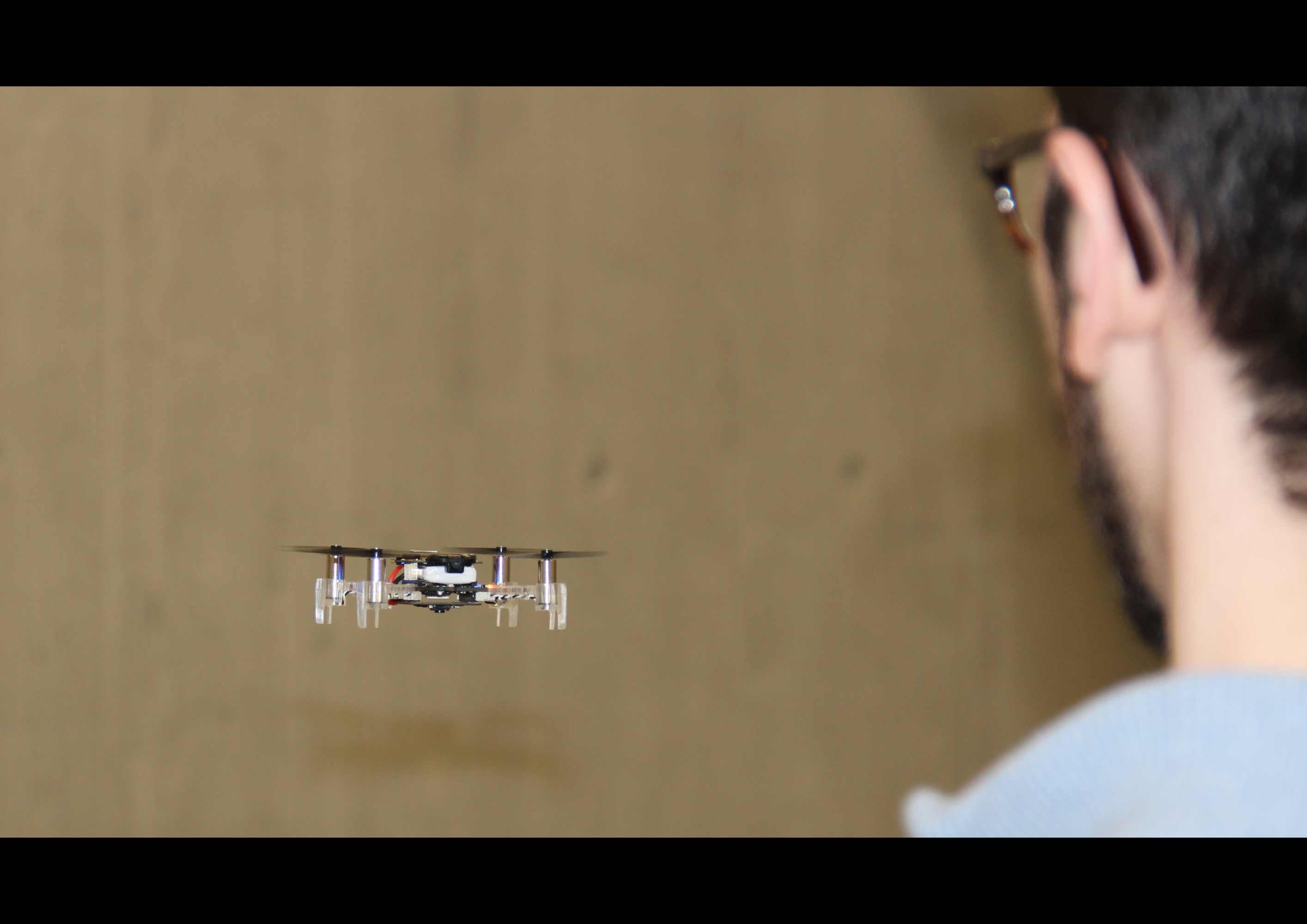}
    \caption{Our prototype based on the COTS Crazyflie 2.1 nano-quadrotor performing the HDI task, with only onboard computational resources.}
    \label{fig:proto}
\end{figure}

This visionary scenario is challenged by the demand for high computational power directly on board, which contrasts with the severe limitations imposed by the nano-UAVs' miniaturized form-factor.
This challenge is further detailed in Table~\ref{tab:taxonomy}, where is presented a taxonomy of the most popular class of vehicles by size~\cite{palossi2017ultra}.
We highlight the approximate order of magnitude for diameter, weight, complete-system power consumption, and the typical class of processors available onboard.
To date, the vast majority of onboard complex robotic perception algorithms have been demonstrated only employing standard/micro-sized UAVs~\cite{trailnet,MAV2018cnnTEGRA}.
These large robots can afford for powerful, but bulky, onboard \textit{embedded computers}, where they can run computationally-intense traditional computer vision (CV) workloads (e.g., based on pattern recognition, feature extraction, simultaneous localization, and mapping) relying on sparse/dense representation of the environment~\cite{SLAM2018Tegra,structure-SLAM2020}.
The price for these approaches is the high demand for computational power and memory, as witnessed by the typical class of devices employed.
For example, the widely-adopted NVIDIA Tegra series~\cite{trailnet,MAV2018cnnTEGRA,SLAM2018Tegra} is capable of tens of \SI{}{\tera Op/\second} within tens of Watts, which is clearly unaffordable on nano/pico-aircraft.
In fact, with a sub-Watt electronic power envelope and a few grams of payload~\cite{Wood2017}, the onboard computing capability of nano-aircraft has been traditionally limited to simple microcontroller units (MCUs).
This class of devices can deliver up to a few hundreds of \SI{}{\mega Op/\second}, resulting insufficient to meet the real-time requirement of state-of-the-art (SoA) perception and navigation algorithms.

A recent research trend demonstrates how addressing the problem at both hardware and algorithmic levels can enable autonomous navigation for the nano-sized class of vehicles~\cite{palossi2019IOTJ,palossi2019DCOSS}.
On the one hand, the combination of the parallel ultra-low-power (PULP) computing paradigm~\cite{pulp} with the heterogeneous architectural model~\cite{conti_hetero} enables flexible and energy-efficient computation within the limited power envelope of a nano-UAV. 
On the other hand, a new class of algorithms based on deep convolutional neural networks (CNNs) represents a lightweight alternative~\cite{mantegazza2019vision,trailnet,MAV2018cnnTEGRA,crash_fly} to traditional perception approaches.

\begin{table}[t]
\caption{UAVs taxonomy by vehicle class-size~\cite{palossi2017ultra}.}
\begin{center}
\begin{tabularx}{\linewidth}{lccc}
\toprule
\small
Vehicle class & $\oslash$ : Weight [cm:kg] & Power [W] & Onboard device\\
\midrule
\text{\textit{standard-size}~\cite{trailnet}}    & $\sim$ 50 : $\geq$ 1		& $\geq$ 100	& Desktop\\ 
\text{\textit{micro-size}~\cite{MAV2018cnnTEGRA}}	& $\sim$ 25 : $\sim$ 0.5 	& $\sim$ 50		& Embedded\\ 
\text{\textit{nano-size}~\cite{palossi2019IOTJ}}	& $\sim$ 10 : $\sim$ 0.01 	& $\sim$ 5		& MCU\\ 
\text{\textit{pico-size}~\cite{Wood2017}}   & $\sim$ 2 : $\leq$ 0.001	& $\sim$ 0.1 	& ULP\\
\bottomrule
\end{tabularx}
\end{center}
\label{tab:taxonomy}
\end{table}

In this work, we address a HDI task that requires a nano-drone to assess its relative pose w.r.t. a free-moving human subject, using low-resolution images acquired from a front-looking camera, as in Figure~\ref{fig:proto}.
The drone's goal is to stay at a constant distance in front of the subject, following their movement.
Our work leverages the PULP paradigm~\cite{palossi2019IOTJ,conti_hetero}, employing, as the onboard processor, a commercial off-the-shelf (COTS) printed circuit board (PCB) from Bitcraze, called \textit{AI-deck}.
This pluggable PCB is compatible with the Crazyflie 2.1 nano-quadrotor and features a PULP-based \textit{GreenWaves Technologies} GAP8 System-on-Chip (SoC) coupled with an ULP QVGA gray-scale image sensor.

Estimating human pose from low-resolution images is a challenging pattern-recognition task. 
In our robot, we solve this problem with a novel streamlined CNN called \textit{PULP-Frontnet}, designed to take advantage of the GAP8 SoC architecture to achieve energy-efficient calculation and precise pose prediction.
Our CNN takes inspiration from the \textit{Proximity} network~\cite{mantegazza2019vision} which addresses the same task, but exploiting a power-unconstrained remote computer fed with high-resolution images, radio-streamed from a standard-size quadrotor.
Our work provides the following contributions beyond the SoA:
\begin{itemize}
    \item we introduce PULP-Frontnet, a novel CNN for pose estimation, which we explore in three variants with different computational, performance, and memory trade-offs on the GAP8 SoC.
    The CNN topologies are designed to meet both the strict power budget of IoT MCUs and the real-time requirement of autonomous nano-drones;
    \item we present our dataset augmentation methodology, which maximizes the model's generalization capability with synthetic pitch, photometric, optical, and geometric enhancements;
    \item using open-source tools~\cite{conti2020nemo,burrello2020dory}, we demonstrate our methodology from perception to control (including training, aggressive 8-bit quantization, CNN deployment, and low-level controller), with no drop in regression performance, even compared to the full precision (float 32-bit) Proximity CNN. 
    We achieve an onboard peak inference performance of \SI{135}{frame/\second} within \SI{86}{\milli\watt} and a top energy efficiency of $\sim$\SI{0.43}{\milli\joule/frame};
    \item we experimentally evaluate how the CNN design impacts on \textit{i}) regression performance, \textit{ii}) power consumption, \textit{iii}) inference rate, and \textit{iv}) closed-loop control accuracy;
    \item we prove our methodology in the field presenting a closed-loop, fully working demonstration of PULP-Frontnet on a 27-grams nano-UAVs, achieving 100\% success-rate on all tests (18 runs on never-seen-before subjects), with behavior comparable with an ideal motion-capture system (median absolute angular error below \ang{5});
\end{itemize}
Our work demonstrates that deep learning models for robotic perception, trained and deployed with the proposed methodology, can afford extreme complexity reduction (up to $24\times$ fewer operations and $33\times$ less memory, vs. the Proximity NN).
Then, offloading our models on an energy-efficient PULP processor, we can achieve a real-time execution even aboard a resource-constrained nano-drone, with no compromise in the quality-of-results, as shown in the supplementary video material.

The rest of the paper is organized as follows: Section~\ref{sec:related_work} provides the SoA overview of deep learning-based nano-UAVs.
Section~\ref{sec:background} introduces the hardware background of our work.
Section~\ref{sec:methodology} presents in detail our \textit{i}) PULP-Frontnet CNN, \textit{ii}) our dataset augmentation methodology, \textit{iii}) the employed training, quantization, and deployment policies, and \textit{iv}) the proposed onboard control.
Section~\ref{sec:results} shows the experimental evaluation of the work, considering \textit{i}) the PULP-Frontnet regression performance, \textit{ii}) the onboard power consumption, inference rate, energy efficiency, and memory use, and \textit{iii}) the final control accuracy with in-field experiments.
Finally, Section~\ref{sec:conclusion} concludes the paper.

\section{Related work}\label{sec:related_work}

For a palm-size ``flying IoT node'', HDI is a first-class use case that can enhance the user experience (e.g., increased safety) thanks to the small size of the node~\cite{hiroi2011size}. 
In this context, our work addresses the pose estimation of a user from low-resolution images, enabling effective HDI.
The \textit{Proximity} NN~\cite{mantegazza2019vision} represents our application baseline, as the same vision-based task is addressed.
This NN is based on ResNet~\cite{resnet2015} and has been demonstrated with a remote commodity desktop computer's GPU.
The Proximity NN is coupled with a Parrot Bebop 2 quadrotor flying near the user and streaming front-looking high-resolution images to the remote computer.
This allows the model to estimate the subject's pose relative to the drone, determine the appropriate control input, and send it back to the drone, achieving its control task, i.e., staying in front of the user. 
Our PULP-Frontnet NN solves the same visual perception task and yields an equivalent quality of robot behavior (see Section~\ref{sec:error_fit}), but employs a novel streamlined DL model, e.g., without residual shortcuts (up to $\sim24\times$ and $\sim33\times$ fewer operations and memory, respectively).
Ultimately, our model runs entirely aboard a Crazyflie 2.1 nano-drone (i.e., around $15\times$ lighter than a Parrot Bebop 2) with no need of any external computer/infrastructure.

Moving into nano-scale UAVs, we focus on those which employ novel deep learning-based algorithms~\cite{palossi2019IOTJ,lowlevelcontrol2019,swarm2019,distance2018,kang2019generalization}.
These approaches are rapidly gaining attention, as they are lightweight compared to the more traditional ones based on the \textit{localization-mapping-planning} cycle~\cite{structure-SLAM2020}, and represent a natural fit within resource-constrained nano-drones.
Among these, we can distinguish two primary ``flavors'': \textit{autonomous} systems that rely only on onboard sensing and computational resources and \textit{automatic} systems that need some off-board aid.

To date, the vast majority of works that combine nano-aircrafts with onboard computation are severely limited on their task's complexity and applicability~\cite{lowlevelcontrol2019, swarm2019}.
In~\cite{lowlevelcontrol2019}, a model-based reinforcement learning (RL) policy is proposed to control a pocket-size quadcopter.
The NN policy runs aboard the drone's MCU, replacing the low-level functionalities provided by a \textit{flight controller} (e.g., control loops), being able to stabilize the drone in hovering for a few seconds.
In~\cite{swarm2019}, the DL paradigm applied to nano-UAVs is further streamlined into the swarm scenario.
The authors introduce a simple DL model composed of two NNs, running in an \textit{inner-outer} fashion, where the inner NN runs as many times as the number of drones in the swarm.
The model predicts the target $z$ component of the drone's pose to keep the fleet's formation during group maneuvers.
It uses a 6-element input vector -- representing relative position and velocity -- per neighboring drone (up to five).
Therefore, their biggest input is between $128-512\times$ smaller than ours, and their peak number of operations to predict $z$ is $\sim$\SI{27}{\kilo MAC}, i.e., three orders of magnitude less than PULP-Frontnet.

Moving to nano-sized \textit{automatic} systems, i.e., requiring off-board resources, the SoA is characterized by a broad spectrum of use-cases and solutions.
For example, vision-based DL/RL algorithms for obstacle avoidance~\cite{distance2018,kang2019generalization,riskaware2017} being computed on external computers, and nano-drones localization systems with additional ad-hoc infrastructure (e.g., ultra-wideband anchors, motion-capture cameras, etc.)~\cite{distance2018, zhao2020learning}.
Offloading computational-intense workloads to remote base-stations can enable complex algorithms fed with abundant sensory data streams from the aircraft.
Nevertheless, this approach has several drawbacks~\cite{SurveyIoT2019Threats}, such as: \textit{i}) latency, \textit{ii}) limited operations distance, \textit{iii}) reliability and security issues on the communication channels, and \textit{iv}) onboard power-consumption overhead due to the high-frequency streaming (e.g., video).

As an alternative to both previous classes of solutions, we are witnessing the advent of ultra-low-power application-specific integrated circuits (ASICs) designed to enable complex functionalities (e.g., visual odometry, simultaneously localization and mapping) aboard small-size robotic platforms~\cite{navion2019,cnnslam2019,slam2020}.
With a power consumption between ten to a few hundreds of \SI{}{\milli\watt}, these systems have been proven compatible with the power envelope of a small-sized UAV.
However, to date, these approaches \textit{i}) have not yet been demonstrated on a real-life flying nano/pico-UAV and \textit{ii}) they only account for one among other fundamental functionalities.
Therefore, ASICs increase the system's complexity because they need co-processors for both basic flying functionalities and to micro-manage control and data transfers. 
On a severely constrained system (i.e., weight and size), this is a big downside and a push towards more integrated solutions (e.g., SoCs).

Lastly, the \textit{PULP-Dronet} project~\cite{palossi2019IOTJ,palossi2019DCOSS} presents an ``halfway'' architectural improvement in the SoA.
Unlike the aforementioned ASIC designs, PULP-Dronet proposes a novel PULP-based hardware design, called \textit{PULP-Shield}, extending the computational capabilities aboard the Bitcraze Crazyflie 2.1 nano-UAV, with a multi-core general-purpose SoC.
This design has been recently launched as a COTS pluggable PCB by Bitcraze, under the commercial name of \textit{AI-deck} -- both the robotic platform and the additional processor are the same adopted in our work.
Additionally, PULP-Dronet presents a vision-based DL algorithm for autonomous driving of a nano-UAV, tackling lane detection and obstacle avoidance tasks.
From a methodology perspective, our work enhances the PULP-Dronet approach in multiple ways:
\begin{itemize}
    \item addressing a 4-output pose estimation task for HDI;
    \item making use of open-source quantization/deployment tools~\cite{conti2020nemo,burrello2020dory}, as well as employing a $2\times$ more aggressive quantization scheme (i.e., 8-bits vs. 16-bits);
    \item including the development flow for ad-hoc dataset collection and its augmentation;
    \item proposing a novel streamlined DL model (up to $10\times$ and $8\times$ fewer operations and memory, respectively);
    \item introducing a thorough model-size analysis to study the relation between power consumption, memory constraints, regression performance, and control accuracy.
\end{itemize}  
Ultimately, our models push further the onboard NN's inference performance with a peak throughput of \SI{135}{frame/\second} @ \SI{86}{\milli\watt} -- PULP-Dronet peaked at \SI{18}{frame/\second} @ \SI{272}{\milli\watt}.

\section{Background}\label{sec:background}

\subsection{PULP paradigm \& GAP8 System-on-Chip}\label{sec:PULP}

Several recently emerged application scenarios (including autonomous flying on nano-UAVs) require local processing capabilities in the order of billions of operations per second on top of devices with power budgets limited to 1--\SI{100}{\milli\watt} -- 10--100$\times$ more than an off-the-shelf microcontroller unit (MCU).
Parallel ultra-low power (PULP) processing is a recently proposed paradigm that is getting industrial and academic traction to respond to this heightened need of performance and energy efficiency for low-power edge devices.
PULP computers couple inside the same System-on-Chip (SoC) a traditional MCU, meant to manage I/O-oriented tasks, with a programmable general-purpose accelerator that is dedicated to the execution of data-parallel computational kernels~\cite{pulp}, such as the linear algebra at the heart of artificial intelligence.

\begin{figure}[t]
    \centering
    \includegraphics[width=\linewidth]{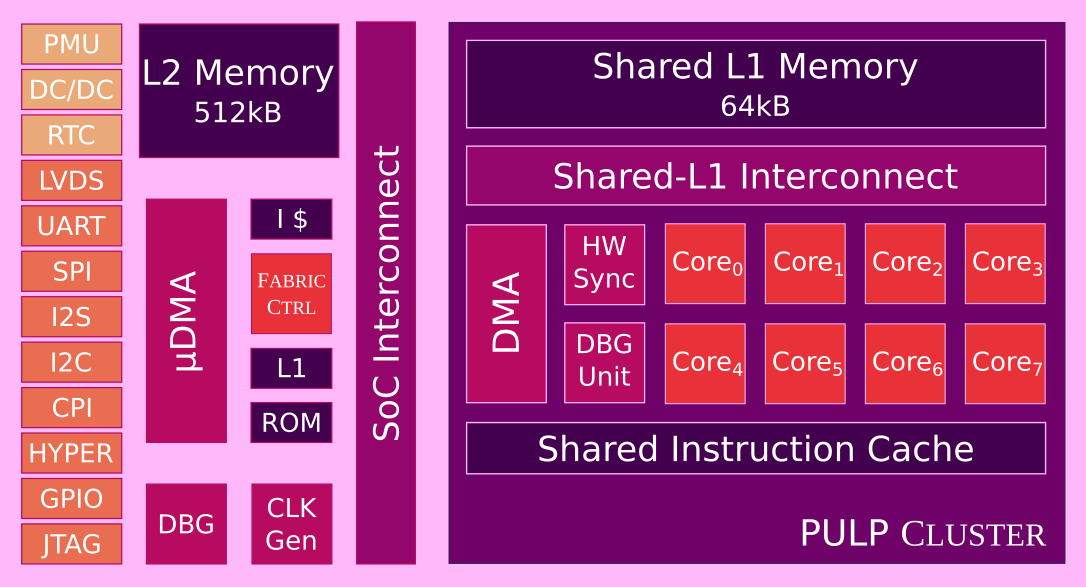}
    \caption{GAP8 System-on-Chip architecture.}
    \label{fig:gap_archi}
\end{figure}

In this work, we focus on a commercial embodiment of the PULP paradigm proposed by GreenWaves Technologies: the GAP8\footnote{https://greenwaves-technologies.com/gap8\_gap9} SoC, shown in Figure~\ref{fig:gap_archi}.
GAP8 employs nine identical RISC-V cores: one, called \textsc{Fabric Controller} (FC), is used as the main core in the MCU; eight are used to build up the parallel general-purpose programmable accelerator, i.e., the \textsc{Cluster} (CL).
All cores are based on the open-source RI5CY design~\cite{GautschinearthresholdRISCVcore2016} and use a relatively simple four-stage in-order single-issue pipeline to implement the \textit{RV32IMC} instruction set.
To improve energy efficiency on integer linear algebra and digital signal processing, RI5CY implements the \textit{XpulpV2} instruction set extension, which includes hardware loops, address post-increment for load/store operations, single instruction multiple data (SIMD) vectorial arithmetic, and dot product instructions for 8- and 16-bit data.

GAP8 includes a full fledged MCU system, organized around the FC.
Data and code are stored on a relatively large (\SI{512}{\kilo\byte}) L2 memory, accessible from both the FC and the CL; the FC has access to a further \SI{16}{\kilo\byte} of private L1 memory (uncached) for data and a \SI{1}{\kilo\byte} instruction cache. 
I/O is performed by means of a programmable controller called micro-DMA~\cite{pullini2017uDMA} ($\mu DMA$), with support for common serial interfaces (e.g., UART, I2C, SPI, I2S) as well as for 8-bit camera parallel interface (CPI).
Additionally, the SoC supports the 8-bit HyberBus protocol by Cypress Semiconductor\footnote{https://www.cypress.com/products/hyperbus-memory}, enabling external L3 DRAMs and Flash memory, with a bandwidth up to \SI{200}{\mega\bit\per\second}.
Each interface is implemented as a separate $\mu DMA$ \textit{channel}; the $\mu DMA$ autonomously moves data between the supported interfaces and the L2 memory, with minimal software intervention.
The channels operate independently one from another, hence different transfers can be partially overlapped in time. 

The main MCU system is accelerated with a \textsc{Cluster} organized around a \SI{64}{\kilo\byte} 16-banks L1 scratchpad memory, with word interleaving.
The L1 is shared by the eight cluster cores through a high-bandwidth interconnect, with zero wait states from all cores in absence of bank collisions.
Program code for both the FC and the \textsc{Cluster} resides in L2 memory.
The 8 cores share a single \SI{4}{\kilo\byte} instruction cache, optimized for a single-program multiple data stream (SPMD) programming model~\cite{loi2018quest}, and use a dedicated hardware block to enable low-latency synchronization.
Data transfers between L2 and cluster L1 are explicit and performed through a programmable DMA controller within the CL domain.

\subsection{Robotic platform \& heterogeneous model}\label{sec:AI-deck}

Our robotic platform is represented by the commercial off-the-shelf (COTS) \textit{Bitcraze Crazyflie 2.1}\footnote{https://www.bitcraze.io/products/crazyflie-2-1} quadrotor, an open-source and open-hardware nano-drone with a weight of \SI{27}{\gram} and a diameter of \SI{10}{\cm}.
The main processor aboard the drone is the STM32F405 microcontroller unit (MCU) that, together with a combined -- i.e., accelerometer/gyroscope -- Bosch BMI088 inertial measurement unit (IMU) lays a reliable ground of basic control functionalities.
The STM32 MCU runs up to \SI{168}{\mega\hertz} and features \SI{192}{\kilo\bit} SRAM and \SI{1}{\mega\bit} flash, on-chip memories, allowing for onboard \textit{inertial state estimation} and \textit{actuation control} tasks.
The former utilizes the IMU's input data to feed an extended Kalman filter (eKF) performing the state estimation at \SI{100}{\hertz}; meanwhile, the latter is embodied by a proportional-integral-derivative (PID) control loop cascade.
The cascade is composed of two control loops, one controlling the attitude at \SI{500}{\hertz}, and a second one updating the position at \SI{100}{\hertz}.

In our configuration, the robotic platform is extended with two COTS pluggable printed circuit board (PCB) from Bitcraze, the \textit{Flow-deck}\footnote{https://www.bitcraze.io/products/flow-deck-v2} and the \textit{AI-deck}\footnote{https://store.bitcraze.io/products/ai-deck}, extending the onboard capabilities.
The former weights \SI{3.5}{\gram} and features the PMW3901 \textit{optical flow} (OF) visual sensor and the VL53L1x time-of-flight (ToF) ranging module.
The OF camera enables the drone to detect its motions in any direction; meanwhile, the ToF sensor provides a distance measurement from the ground.
This two sensory information is forwarded to the onboard state estimation increasing its accuracy and reliability, for example, reducing long-term drift.
The second expansion board, the AI-deck, represents the high-level onboard computing device in charge of executing complex -- otherwise non-addressable -- navigational algorithms such as the proposed CNNs.

\begin{figure*}[t]
	\centering
	\includegraphics[width=\textwidth]{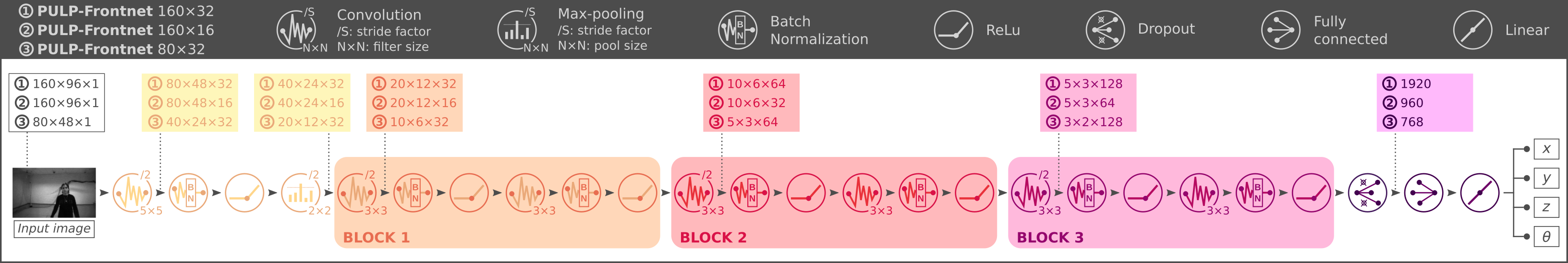}
	\caption{PULP-Frontnet neural network, exploring three model sizes, varying memory and computational requirements.}
	\label{fig:pulp_frontnet}
\end{figure*}

The AI-deck weights \SI{4.4}{\gram} and is the first commercial embodiment of the \textit{visual navigation engine} -- called PULP-Shield -- introduced in~\cite{palossi2019IOTJ,palossi2019DCOSS}.
Like its predecessor, the AI-deck features a general-purpose GAP8 SoC, additional off-chip memories as big as \SI{512}{\mega\bit} HyperFlash and \SI{64}{\mega\bit} HyperRAM, and a Himax HM01B0 ULP monochrome QVGA camera.
The only exception w.r.t. the first PULP-Shield prototype is represented by the additional ESP32-based WiFi\footnote{https://www.u-blox.com/en/product/nina-w10-series-open-cpu} transceiver and a UART communication channel between the STM32 and the GAP8, instead of the SPI one initially proposed.
Even if the availability of the WiFi module eases remote visual-based computation, in the rest of this work, we will refer to a configuration where the expensive WiFi (up to multiples order of magnitude higher power consumption than the SoC) is always turned off.
Our primary mission is to develop a fully autonomous system where the whole navigation intelligence runs aboard the nano-drone without any external communication/computation.
In our case, the only exception of active WiFi transmission is for dataset collection and showcasing purposes.

The combination of the STM32 MCU with the GAP8 embodies the \textit{host-accelerator} heterogeneous model at the ULP-scale~\cite{conti_hetero}.
The \textit{host} (i.e., STM32) is devoted to control-oriented tasks and part of the sensor interfacing.
Instead, the computational intensive navigation workloads are offloaded to the general-purpose \textit{accelerator} (i.e., GAP8).
Due to the AI-deck's multi-level (L1-L2-L3) memory hierarchy and the pairing of the low-resolution camera with the SoC, the system minimizes communication overhead (e.g., input images are directly fed to the GAP8 without any need to pass through the host) and exploits the locality of data.
Even if these basic concepts and functionalities are designed for nano-drones autonomous driving scenario, they are general and applicable to any IoT node requiring visual processing capabilities.

\section{Methods}\label{sec:methodology}

\subsection{PULP-Frontnet neural network}\label{sec:pulp-frontnet}

To solve our pose estimation task, we present the \textit{PULP-Frontnet} CNN, shown in Figure~\ref{fig:pulp_frontnet}.
The proposed neural network is inspired by the original \textit{Proximity} network~\cite{mantegazza2019vision}, where the same task was addressed with different ResNet-based topology and robotic platform.
Our model takes as input a front-looking gray-scale image from the low-resolution camera aboard the nano-drone and outputs four independent variables, defining the target pose.

In our design, we employ a classical pattern where each convolution is followed, in order, by a batch normalization and activation (i.e., Relu) stage, stabilizing the learning process with a per-layer scaling effect~\cite{ioffe2015batch}.
The model is characterized by a first $5\times5$ convolutional layer, followed by a $2\times2$ max-pooling.
Each of them reduces by $4\times$ the output feature map size due to a striding factor of two (both horizontal and vertical).
Then a \textit{block pattern} of two $3\times3$ convolutional layers is repeated three times, where each block doubles the number of output channels and reduces by $4\times$ the output feature map size.
The last part of the model presents a dropout stage followed by a fully connected layer that outputs the pose as a point in the 3-dimensional space (\textit{x, y, z}), and a rotation angle w.r.t. the gravity z-axis ($\theta$). 

To successfully deploy PULP-Frontnet on top of a resource-constrained MCU, such as the GAP8 SoC, the NN's execution must comply with the strict real-time constraints dictated by the application scenario while respecting the bounds imposed by the on-chip and onboard resources.
The main constraints can be summarized as follow: \textit{i}) \textit{throughput or minimum frame-rate}, \textit{ii}) \textit{quality-of-result or regression performance}, and \textit{iii}) \textit{onboard/on-chip memory limits}.
In this light, it is clear we need a reliable methodology and strategy to reduce the memory and computational loads, to ease the deployment on the available resources while exploiting the hardware architecture at best to meet the real-time constraint.

Therefore, to comply with the given architectural constraints, we introduce fixed-point arithmetic and 8-bit integer data (see Section~\ref{sec:quantization}) instead of floating-point calculation on a 32-bit data type -- as per the original proximity network.
This transformation represents an industry-standard with many advantages, such as 4$\times$ reduction in the memory need; fast and efficient execution on devices without hardware support for floating-point calculation, such as the GAP8 SoC and many other commercial MCUs; and enabling for optimized signal processing instructions (e.g., packed-SIMD). 
The price for these enhancements at both memory and computational level is a minimal drop in the CNN's accuracy~\cite{choi2018pact}.

\begin{figure*}[t]
	\centering
	\includegraphics[width=\textwidth]{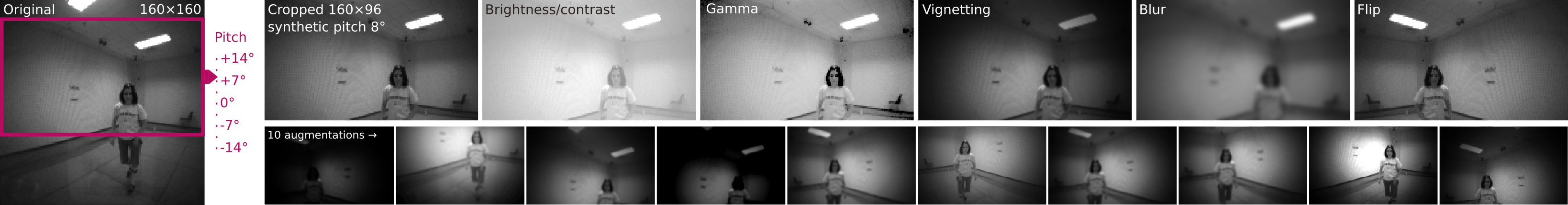}
	\caption{The original dataset image (left) is cropped at a random height to simulate pitch variations; a random subset of photometric, optical and geometric augmentations (top) are then applied. Bottom: ten random augmentations originating from the same source image.}
	\label{fig:dataset}
\end{figure*}

For this reason, in this work, we aim at investigating the relationship between memory footprint and computational requirements (i.e., number of operations) of the proposed model w.r.t. its regression performance and closed-loop in-field control accuracy (see Section~\ref{sec:error_fit} and~\ref{sec:in-field}).
Therefore, we play with the memory/operations knobs by varying input image size and the number of channels between different NN's blocks, affecting both parameters (i.e., weights) and intermediate feature map sizes.
This process results in three PULP-Frontnet NN variants, as shown in Figure~\ref{fig:pulp_frontnet} and detailed in Table~\ref{table:NN_requirements}.
Here, the number of operations accounts only for convolutional and fully connected layers; instead, the memory requirements consider the input image, all weights, and all intermediate buffers to store the feature maps -- i.e., what we would obtain from a straightforward implementation.

The first version, named 160$\times$32, is characterized by the biggest memory footprint and the highest number of multiply-and-accumulate (MAC) operations needed to perform one input image inference.
The 160$\times$16 NN represents the extreme edge on the memory minimization exploration axis instead.
Lastly, the NN called 80$\times$32 uses a smaller input image, w.r.t. the previous two NNs, showing the minimum computational requirements.
Table~\ref{table:NN_requirements} also compares the proposed PULP-Frontnet models with the Proximity NN.
With 7-24$\times$ fewer operations and 12-33$\times$ less memory need, if compared to the Proximity NN, our model makes it possible to envision an outstanding performance on our deployment robotic platform.

\begin{table}[h]
\caption{MAC operations and memory footprint for one frame inference of the PULP-Frontnet models and the Proximity NN~\cite{mantegazza2019vision}.}
\begin{center}
\begin{tabularx}{\linewidth}{lcccc}
\toprule
\small
\textit{PULP-Frontnet} & $160 \times 32$ & $160 \times 16$ &\multicolumn{1}{c|}{$80 \times 32$} & NN~\cite{mantegazza2019vision}\\
\midrule
Operations [MMAC] & 14.1 & 4.3 & 4.0 & 96.5\\
Memory [\SI{}{\kilo\byte}] & 499 & 184 & 348 & 6116\\
\# Parameters & 3.03$\times10^5$ & 7.80$\times10^4$ & 2.99$\times10^5$ & 1,26$\times10^6$\\
\bottomrule
\end{tabularx}
\end{center}
\label{table:NN_requirements}
\end{table}

\subsection{Dataset collection \& augmentation}\label{sec:dataset}

\textbf{Dataset collection.}
The dataset used to train, validate, and test the PULP-Frontnet models is collected in a 10 $\times$ \SI{10}{\meter} room equipped with a motion capture system (mocap), composed of 12 Optitrack PM13 cameras.
The dataset is acquired using the same deployment robotic platform, introduced in Section~\ref{sec:AI-deck}, and therefore using the onboard QVGA, gray-scale, Himax camera.
In our dataset collection setup, the quadrotor is equipped with a mocap target (i.e., reflective marker) and affixed with a horizontal attitude (zero pitch and roll) on a wheeled cart with adjustable height. 

During the dataset acquisition, an operator moves the cart around the room, continuously changing its position, heading, and height (in a range between and \SIrange{1.20}{1.45}{\meter}).
Simultaneously, the recorded subject moves freely in the environment, wearing either a baseball cap or an almost-invisible headband with a mocap target affixed.
The operator and the subject move so that the latter is visible in most frames while capturing a wide distribution of camera-subject distances and headings.
Because the camera is moving and the room's setup (e.g., lighting) is purposefully changed between different recording sessions, the images contain varied backgrounds, sometimes cluttered with objects including furniture, lab equipment, and people other than the subject.

Thanks to the markers applied, on both quadrotor and subject, the mocap system can track their pose at \SI{200}{\hertz}, resulting in a precise ground truth labeling information.
These poses are recorded using the ROS~\cite{quigley2009ros} framework and synchronized with the $160 \times 160$ pixels video frames (streamed to a host computer via WiFi from the quadrotor). Once the quadrotor and subject poses are known in the room reference frame, the relative pose of the subject with respect to the drone is computed and decomposed in its $x$, $y$, $z$ and $\theta$ components.
We recorded ten distinct sessions featuring different subjects of various ages, height, clothing, and hairstyles.
Furthermore, some subjects wear face masks and eyeglasses only for part of the session to enhance data variety.
In total, we collected 6657 frames representing 25 minutes of acquisition time (\SI{150}{\second} per subject at $\sim$\SI{4.5}{frame/\second}).

Among the ten recorded sessions, we use six for training and the remaining four for evaluation; this ensures that the same subject does not appear both in training and evaluation sets.
Therefore, our evaluation metrics quantify the model's ability to generalize to unseen subjects and do not reward overfitting on the training ones.
The training dataset is further split by holding out a random 20\% of its frames for validation, for which the loss is monitored as training progresses.
The model that yields the lowest validation loss is selected following a standard \textit{early-stopping} pattern.

\textbf{Dataset augmentation.}
During a mission, the quadrotor pitch changes in a range of approximately $\pm \SI{15}{\degree}$ to generate forward/backward accelerations: this heavily affects the image acquired, which is not stabilized optically nor electronically.
Therefore, a CNN trained only on images acquired with a flat attitude returns increasingly inaccurate results as the in-field acquisition pitch deviates from $\pm$\SI{0}{\degree}.

Our mitigation strategy to this problematic effect is to apply, during training, a synthetic pitch augmentation technique (see Figure~\ref{fig:dataset}) to ensure that our model is robust to the actual pitch of the quadrotor during the mission.
In particular, we observe that we can generate $160\times96$ pixel images with an approximated pitch in the range $\pm \SI{14}{\degree}$ by cropping a subset of the rows from a $160\times160$ pixel image acquired with a horizontal and constant attitude.
For example, the top 96 rows of a dataset image, compared to its middle 96 rows, depict the same scene with an approximated pitch of $+$\SI{14}{\degree} (see Figure~\ref{fig:dataset} on the left).
This approximation disregards photometric effects due to vignetting in the full-frame image and ignores perspective distortion; however, the approximation is acceptable for our purposes and can be implemented as an inexpensive augmentation strategy within the training pipeline. 
In particular, for a given $160\times160$ pixel dataset image, we crop $160\times96$ training samples at random vertical positions; this yields more than 66'000 training instances.

To further promote generalization, we apply additional data augmentation techniques, as shown in Figure~\ref{fig:dataset}. 
First, we apply the following photometric and optical augmentations, each independently sampled with a 50\% probability:
\begin{itemize}
    \item \textit{contrast} jittering (multiplicative factor uniformly sampled in the range $[0.7, 2.0]$), accounting for erratic auto exposure behavior in our camera model;
    \item \textit{brightness} jittering in the range $[-0.2, 0.2]$;
    \item \textit{gamma} correction (exponent uniformly sampled in the range $[0.4, 2.0]$);
    \item synthetic \textit{vignetting} effect with random radius and strength, which accounts for the strong vignetting present in the Himax images, and its likely variability in different cameras and illumination conditions;
    \item smoothing using a gaussian kernel with $\sigma = 3$ pixels, accounting for \textit{blurring} due to camera motion, vibration, and lens defocus.
\end{itemize}
Finally, with a 50\% probability, we horizontally \textit{flip} the image while correspondingly altering the ground truth, i.e., negate the $y$ and $\theta$ variables. 
Note that this automatically ensures that the distribution of these variables in the datasets is symmetric.

\subsection{Training, quantization, and deployment.}\label{sec:quantization}

\textbf{Network training and quantization.}
We divide the procedure to produce a deployable PULP-Frontnet in several steps, implemented using the PyTorch framework and the open-source NEMO library~\cite{conti2020nemo}.
First, we train a \textit{full-precision} floating-point PULP-Frontnet on the dataset described in Section~\ref{sec:dataset}, minimizing the L1 loss for the pose vector ($x,y,z,\theta$).
We use the Adam optimizer with learning rate $10^{-4}$ and early stopping over 100 epochs; we then select the model with the lowest validation loss, which is obtained after on average 80 epochs.

After full-precision training has ended, we perform a \textit{fake-quantized fine-tuning} step.
We use linear uniform per-layer quantization and a variant of PACT~\cite{choi2018pact} for training, with quantization functions of the form:
\begin{equation}
    Q(\mathbf{t}) = \varepsilon_\mathbf{t}\cdot \left\lfloor \frac{\mathbf{t}}{\varepsilon_\mathbf{t}} \right\rfloor
    \label{eq:quant}
\end{equation}
where $\mathbf{t}$ is the tensor, $Q(\mathbf{t})$ its quantized representation, and $\mathbf{\varepsilon_t}$ is a scalar representing the difference between two consecutive fixed-point values.
We chose to use 256 levels (8 bits) for activations and 128 levels for weights; given the weights' asymmetric distribution around 0, these 128 levels can be represented using an 8-bit signed integer.
The network is manipulated so that convolutional and fully connected layers use weights that have passed through the quantization function of Equation~\ref{eq:quant}, with $\varepsilon_\mathbf{W} = (\mathbf{W}_\mathrm{max}-\mathbf{W}_\mathrm{min}) / (2^{7}-1)$.
$\mathbf{W}_\mathrm{max}$ and $\mathbf{W}_\mathrm{min}$ are fixed to the layer-wise maximum and minimum values of weights, respectively.
Batch normalization and pooling layers are left untouched at this stage.

To quantize activations, we replace all ReLUs in the network with the quantization function of Equation~\ref{eq:quant}, choosing $\varepsilon_\mathbf{x} = \alpha / (2^{8}-1)$.
$\alpha$ is initialized to the maximum value reached by the output of each ReLU over the validation set and is then trained by back-propagation.
To fine-tune the fake-quantized network, we initialize it with the equivalent quantized value of the full-precision weights, then we calibrate the $\alpha$ parameters using the validation set, and finally, we perform 100 epochs using the Adam optimizer to minimize the L1 loss.
We use an initial learning rate of $10^{-4}$ with 0.95 decay and weight decay of $10^{-6}$.
After the fine-tuning procedure has completed, the network is transformed into an \textit{integer deployable} one~\cite{conti2020nemo}.
Weights $\mathbf{W}$ are approximated without network performance loss as:
\begin{equation}
    \mathbf{W} \approx\varepsilon_\mathbf{W}\cdot \mathbf{W^*}_\mathrm{min} + \varepsilon_\mathbf{W}\cdot {\mathbf{W^*}} \label{eq:1}\;, 
\end{equation}
where $\mathbf{W^*}_\mathrm{min}=Q(\mathbf{W}_\mathrm{min}) / \varepsilon_\mathbf{W}$ and $\mathbf{W^*}$ is an integer tensor with values in the range $[-64,+63]$.
Notice that $\mathbf{W^*}_\mathrm{min}$ is also defined in the same $[-64,+63]$ region; therefore, $\mathbf{W}$ as a whole can be accurately represented with an 8-bit signed integer even if weights are distributed asymmetrically around 0. 
We replace floating-point batch normalization layers with integer ones using 32-bit parameters.
Finally, tensors outcoming activation layers are represented using 8-bit unsigned integers, while intermediate outputs use a 32 bits data type.
Therefore, the entire network can run entirely in the integer domain and produces a vector of four 32-bit fixed-point values that approximate ($x,y,z,\theta$).

\textbf{Deployment strategy.}
The quantized PULP-Frontnet models' deployment is based on the PULP-NN library~\cite{burrello2020dory} for optimized 8-bits fixed-point arithmetic.
PULP-NN exploits the eight cores general-purpose \textsc{Cluster} of the GAP8 SoC to parallelize kernels' execution on the spatial dimensions (i.e., $width \times height$) and the SIMD and bit-manipulation ISA extensions to achieve the best performance and energy efficiency, peaking at \SI{15.6}{MAC/cycle} for squared-size images.
However, the available kernels operate on the shared L1 \SI{64}{\kilo\byte} memory, constraining their applicability to small layers; therefore, these kernels are not suitable for deploying our network without any additional intermediate manipulation.

For this purpose, we employ the DORY tool~\cite{burrello2020dory} that automatically produces C ``wrapping'' code to manage the two levels of on-chip memory (i.e., L1, L2) and the external RAM, orchestrating weights and activation movements to maximize PULP-NN kernels performance.
Thanks to general templates and tensor tiling, DORY divides the layers in nodes, which are executed in L1 by inserting \textit{i}) double-buffered L2-L1 DMA calls, and \textit{ii)} calls to basic kernels in the PULP-NN library, which performs computation on local L1 data.
The data movements are always overlapped with computation due to asynchronous and non-blocking DMA calls. 
Further, DORY always operates storing the network's weights in the external RAM, loading them into the L2 during the previous layer's execution by employing the $\mu$DMA -- i.e., during the execution of layer \textit{i}, the weights of layer \textit{i+1} are transferred --, realizing a two-level double buffering between RAM, L2, and L1.

This policy enables the execution of NNs whose weights would not fit in the available L2 memory constraint. 
However, it prevents the possibility of ``ahead-of-time'' weights pre-loading into the L2 for small networks that would not benefit from this continuous data movement.
Therefore, for our exploration on the relation between memory, power consumption, and the CNN regression performance, we manually modify the C code produced by DORY to investigate our smallest PULP-Frontnet model (i.e., 160$\times$16) pre-loading all the weights in L2.
In this way, we restrict the RAM utilization only to an initialization stage before the mission starts.

\subsection{Onboard closed-loop control}\label{sec:control}

\begin{figure}[t]
    \centering
    \includegraphics[width=\linewidth]{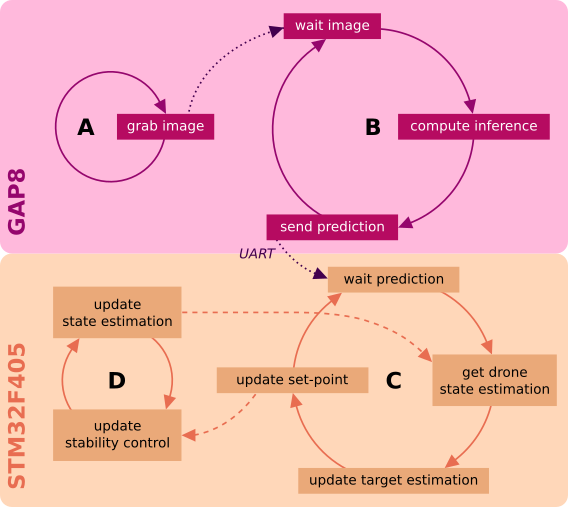}
    \caption{The four main loops that define to the drone behavior. (A) \textit{camera loop} and (B) \textit{inference loop} run on the GAP8 SoC, while (C) \textit{high-level control loop} and (D) \textit{low-level control loop} run on the STM32F405. Dark violet dotted arrows mean synchronization.}
    \label{fig:control_loops}
\end{figure}

Figure~\ref{fig:control_loops} illustrates the control flow that makes the drone hover in front of a person.
Pose estimations are computed on the GAP8 SoC by the PULP-Frontnet CNN and then sent through the UART interface to the Crazyflie's flight controller (STM32F405 MCU), where they are filtered and then used to compute low-level control set-points.
This flow is organized in four loops:
\begin{description}[leftmargin=!,labelwidth=10pt]
    \item[(A) camera loop:] the camera regularly grabs 162$\times$162 gray-scale images;
    \item[(B) inference loop:] once a new image is available, it is cropped to the target size (e.g., 160$\times$96) and pushed to the inference loop. The predicted pose in the drone-relative frame is sent through UART from the GAP8 to the STM32; 
    \item[(C) high-level control loop:] the STM32 receives the pose and transforms it to the drone odometry frame, fused with the previous target state estimation, and uses it to update a low level set-point; 
    \item[(D) low-level control loop:] the high-rate stability loop regularly updates the drone's state estimation and applies a cascade of PID controllers to reach the set-point decided by the high-level loop.
\end{description}
The only synchronization points between different loops are \textit{i}) on the wait for a new image in the inference loop, and \textit{ii}) on the wait for a new pose in the high-level control loop.
Therefore, the high-level control's frequency is bounded by the maximum inference loop (i.e., up to \SI{135}{\hertz} for the fastest 80$\times$32 NN), which in turn also depends on the maximum image grabbing rate (i.e., up to $\sim$\SI{160}{\hertz}).

\textbf{Notation and state space.}
As presented in Section~\ref{sec:dataset}, PULP-Frontnet outputs a pose estimation that does not depend on the pitch and roll components of the orientations of the drone and the human subject.
Therefore, we only consider poses that belong to the Euclidean subgroup $\mathrm{SE}(3)$ generated by translations and rotations around the common z-axis aligned with gravity.
We denote the pose of object $\mathcal{A}$ -- drone or subject -- with respect to frame $\mathcal{B}$ as $p^{\mathcal{B}}_{\mathcal{A}} = (\vec p, \theta) \in \mathbb{R}^3 \times S^1 $, where $\vec p$ represents the position and $\theta$ the rotation around the common z-axis.
To better react to the subject's movements, the drone also keeps track of their linear and angular velocity $v$ as part of the subject's state.
We denote the state of object $\mathcal{A}$ with respect to frame $\mathcal{B}$ as 
$\xi^{\mathcal{B}}_{\mathcal{A}} = (p^{\mathcal{B}}_{\mathcal{A}}, v^{\mathcal{B}}_{\mathcal{A}}) \in \mathbb{R}^3 \times S^1 \times \mathbb{R}^4$.

\begin{figure}[t]
    \centering
    \includegraphics[width=\columnwidth]{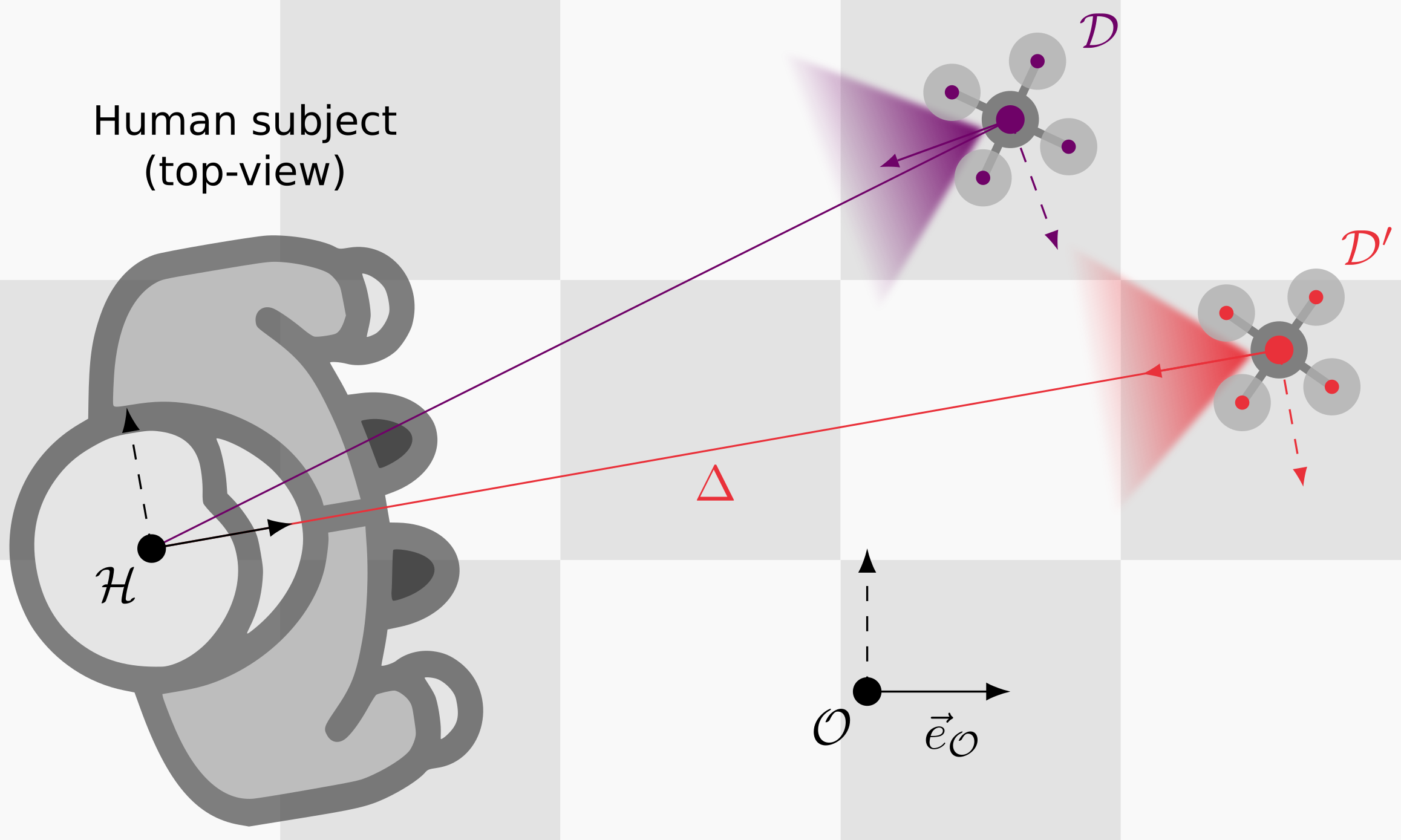}
    \caption{Our three reference frames: $\mathcal{D}$, $\mathcal{H}$, and$\mathcal{O}$. Top-down view of the human subject $\mathcal{H}$ walking sideways to their right and the drone $\mathcal{D}$ (violet) trying to stay in front at distance $\Delta$ by moving towards target pose $\mathcal{D}'$ (red).}
    \label{fig:control}
\end{figure}

We introduce three frames, which all share the same z-axis orientation: $\mathcal{D}$ attached to the drone, $\mathcal{H}$ attached to the subject, and $\mathcal{O}$ as the world-fixed drone odometry frame.
In Figure~\ref{fig:control}, we depict the top-down view of the human subject $\mathcal{H}$ walking sideways to their right and the drone $\mathcal{D}$ (violet) trying to stay in front at constant distance $\Delta$.
It does so by moving towards target pose $\mathcal{D}'$ (red) while rotating towards the violet line. 
Frames are drawn with a solid x-axis (with a unit vector $\vec e$ that points away from the front of the object), a dashed y-axis, and share the same z-axis exiting the drawing.
All computations, except inference, are done in the odometry frame $\mathcal{O}$; therefore, we later drop the related index to simplify the notation.
In the following, differences between angles in $S^1$ are meant as real values in $[-\pi, \pi]$. 

\textbf{High-level control loop.}
As illustrated in loop C of Figure~\ref{fig:control_loops}, the high-level control loop, which is in charge of updating the low-level set-point, comprises four steps:
\begin{itemize}[leftmargin=!,labelwidth=10pt]
    \item \textit{wait prediction:} the loop waits until a new prediction $\tilde p_\mathcal{H}^\mathcal{D}$ is computed from the inference loop, based on the current camera image;
    \item \textit{get drone state estimate:} the loop reads and stores the current drone state estimation $\xi^{\mathcal{O}}_{\mathcal{D}}$ from the low-level control;
    \item \textit{update target estimation:} using the current transformation $\xi^{\mathcal{O}}_{\mathcal{D}}$, the loop computes the prediction to the odometry frame $\tilde p_\mathcal{H}^\mathcal{O}$, which it uses to update the subject state estimation $\xi_\mathcal{H}^\mathcal{O}$ in a Kalman filter;
    \item \textit{update set-point:} finally, the loop updates the desired velocity $v_\mathcal{D}^\mathcal{O}$ to make the drone staying in front of the subject, the desired distance $\Delta$.
\end{itemize}

\textbf{Kalman filter.}
We model the drone dynamics and inference prediction as a stochastic linear process with normally-distributed and zero-mean noise. 
More precisely, we let $(p_n, v_n)$ be the values of $\xi_\mathcal{H}^\mathcal{O}$ and $o_n$ those of $\tilde p_\mathcal{H}^\mathcal{O}$ at time $t_n$, and assume that:
\begin{eqnarray}
p_{n+1} &=& p_{n} + v_n (t_{n+1} - t_n) \\
v_{n+1} &=& v_{n} + a (t_{n+1} - t_n) \\
o_{n} &=& p_{n} + \epsilon,
\end{eqnarray}
where the acceleration $a \in \mathbb{R}^4$ has covariance $Q$ and zero mean, and the observation error $ \epsilon \in \mathbb{R}^3 \times S^1$ has covariance $R$ and zero mean.
We make two further assumptions: the processes are isotropic, and invariants, i.e., covariances $Q$ and $R$ are constant and diagonal.
We can then decouple the Kalman filters for each component at the cost of neglecting that predictions, w.r.t. the drone longitudinal and lateral axis, have slightly different MSE and that errors depend on the human subject relative position.

\textbf{Velocity control.}
We define simple, uncoupled linear and angular controls to let the drone hover in front of the human subject.
The target linear velocity $\vec{v}'_{\mathcal{D}}$ is computed to allow the drone to reach the target position $\vec p_{\mathcal{D}'} $ at distance $\Delta$ in front of the subject in time $\tau$.
Assuming the subject is smoothly moving at an almost constant speed, 
we add to  $\tau$ the subject's estimated velocity $\vec v_\mathcal{H}$, where:
\begin{eqnarray}
\vec p_{\mathcal{D}'} &=& \vec p_{\mathcal{H}} + \vec e_{\mathcal{H}} \Delta \\
\vec{v}'_\mathcal{D} &=& \left. \frac{\vec p_{\mathcal{D}'} - \vec{p}_\mathcal{D}}{\tau} + \vec v_\mathcal{H} \right |_{[-\vec v_{\max}, \vec v_{\max}]}.
\end{eqnarray}
The goal of the angular control is to keep the subject centered in the image frame.
Therefore, we first compute a target orientation $\theta'_\mathcal{D}$ as the current orientation to face the person and then the angular speed to reach it over time $\tau$:
\begin{eqnarray}
\theta'_\mathcal{D} &=& \angle (\vec{e}_\mathcal{D}, \vec{p}_\mathcal{H} - \vec p_\mathcal{D}) \\
\omega'_\mathcal{D} &=& \left. \frac{ \theta'_\mathcal{D} - \theta_\mathcal{D}}{\tau} \right |_{[- \omega_{\max}, \omega_{\max}]}.
\end{eqnarray}
All velocities are clamped within the maximal ranges of $v_{\max}=\SI{1}{\meter\per\second}$ for the linear speed, and the angular speed below $\omega_{\max}=\SI{0.8}{\radian\per\second}$.

\textbf{Low-level control loop.}
The lowest level of our control is based on the open-source \texttt{Controller\_PID} offered by the Crazyflie 2.1 firmware\footnote{https://github.com/bitcraze/crazyflie-firmware}.
In particular, the drone applies a cascade of PID controllers (with no synchronization with the high-level controller) to update \textit{i}) the target attitude from the current speed and target velocity (@\SI{100}{\hertz}), \textit{ii}) the target attitude rate from the current attitude and the target attitude (@\SI{500}{\hertz}), and \textit{iii}) motor commands from the current attitude rate and the target attitude rate (@\SI{500}{\hertz}).
We limit the absolute value of the target pitch to \SI{12}{\degree} so to respect the pitch limit used for the dataset augmentation.
The low-level control loop is also in charge of updating the drone state estimation $\xi^{\mathcal{O}}_{\mathcal{D}}$, fusing the onboard measurements from the IMU, OF camera, and ToF distance sensor (z-direction) using an extended Kalman filter.

\section{Results}\label{sec:results}

In Section~\ref{sec:error_fit}, we evaluate the regression performance of different models with offline experiments on the testing set.
In particular, we compare full precision and quantized variants of the same network, different network architectures, and quantitative performance across the output variables.
Section~\ref{sec:onboard_perf} evaluates energy efficiency, power consumption,  and computational/memory requirements of the proposed PULP-Frontnet variants.
Finally, Section~\ref{sec:in-field} analyzes the quadrotor behavior on in-field person-tracking experiments, using only onboard sensing and computation.

\subsection{Regression performance}\label{sec:error_fit}

In this section, we report the regression performance metrics for our three PULP-Frontnet networks (both full-precision and quantized) and for the Proximity NN~\cite{mantegazza2019vision} (only full-precision).
All models are trained on the same training set and augmented as defined in~\ref{sec:dataset}.
Inputs are scaled to the appropriate input resolution for each model ($160\times96$ and $80\times48$ for the PULP-Frontnet, and $108\times60$ for Proximity NN) using bilinear interpolation.
All models are then evaluated on the same testing set, sampled without augmentation ($\sim$4'000 images).

\begin{table}[h]
\caption{Regression performance for our networks (quantized) and Proximity NN~\cite{mantegazza2019vision} (full precision).}
\begin{center}
\begin{tabularx}{\linewidth}{lXXXXXXXX}
\toprule
\small
\multirow{2}{*}[-0.5\dimexpr \aboverulesep + \belowrulesep + \cmidrulewidth]{Network} & \multicolumn{4}{c}{MAE [$\cdot 10^{-3}$]} & \multicolumn{4}{c}{MSE [$\cdot 10^{-3}$]} \\
\cmidrule(lr){2-5}\cmidrule(l){6-9}
                &  $x$ &  $y$ &  $z$ & $\theta$ &  $x$ &  $y$ &  $z$ & $\theta$ \\
\midrule
$160\times32$   & \textbf{195} & 192 & 101 & 482 & \textbf{66} & \textbf{78} & \textbf{20} & 386\\
$160\times16$   & 203 & 191 & 110 & 492 & 74 & 83 & 25 & 412\\
$80\times32$    & 226 & \textbf{178} & 106 & 556 & 88 & 84 & 29 & 504\\
NN~\cite{mantegazza2019vision} & 210 & 219 & \textbf{96} & \textbf{473} & 79 & 91 & 21 & \textbf{385}\\
\bottomrule
\end{tabularx}
\end{center}
\label{tab:static_performance}
\end{table}

\begin{figure*}[t!]
	\centering
	\includegraphics[width=\textwidth]{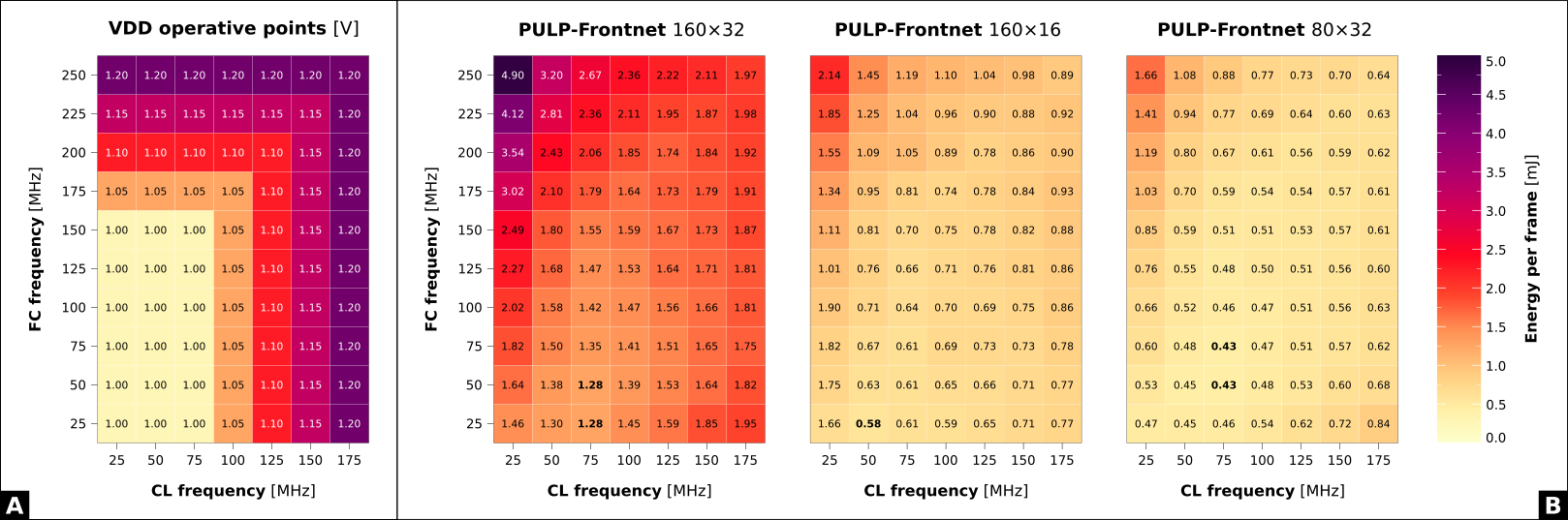}
	\caption{\textit{A}) VDD operative points used for the frequencies sweeping in B. \textit{B}) Energy efficiency analysis, as the energy required to run the inference on a single frame, for all three NNs, sweeping both FC and CL frequency. The most energy-efficient operative points are reported in \textbf{bold} (per NN).}
	\label{fig:energy}
\end{figure*}

Table~\ref{tab:static_performance} reports, for each output variable, the mean absolute error (MAE), expressed in meters for $x,y,z$ and radians for $\theta$, and the mean squared error (MSE), highlighting the best score (in bold) for both metrics across all the networks.
The MAE on $x$ and $y$ variables is $\sim$\SI{0.2}{\meter} for all models, which indicates a good ability to localize the subject on the horizontal plane.
For the $z$, the error is lower w.r.t. other variables due to its reduced variance.
Compared to MAE, MSE values further penalize large errors: on this metric, the $160\times32$ model consistently outperforms smaller networks, which is explained by its utilization in training.
Additionally, Table~\ref{tab:static_performance} shows how the proposed PULP-Frontnet models have comparable performance with the original Proximity network, which has many more parameters and higher computational/memory requirements (see Section~\ref{sec:pulp-frontnet} and Table~\ref{table:NN_requirements}).

\begin{figure}[b!]
    \centering
    \includegraphics[width=\linewidth]{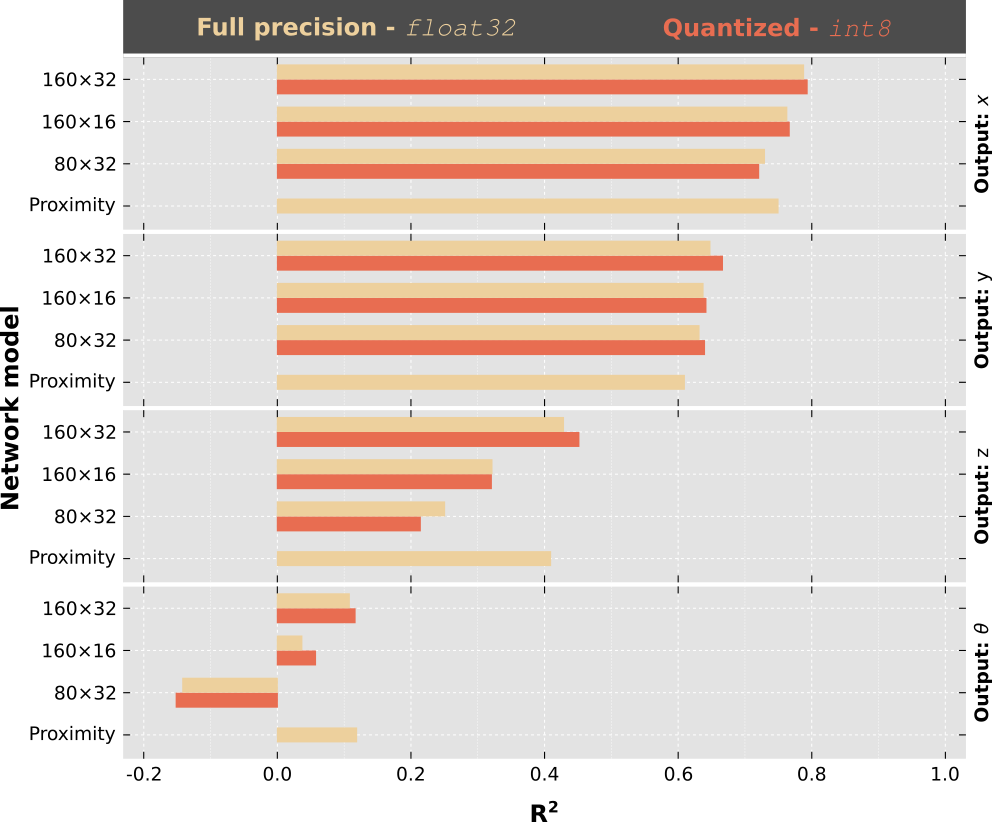}
    \caption{Regression performance ($R^2$) on the testing dataset.}
    \label{fig:R2}
\end{figure}

Figure~\ref{fig:R2} presents a comparison based on the coefficient of determination $R^2$, a standard adimensional metric for regression performance.
$R^2$ represents the fraction of the variance in the target variable explained by the model (higher is better).
A trivial model that predicts the average of the target variable for the whole testing dataset yields $R^2 = 0$; in this case, the MSE is the same as the target variable's variance -- i.e., the model captures no useful information other than the average.
At the other extreme, an ideal model yields $R^2 = 1.0$, and lastly, if $0 < R^2 < 1$, the model can account for only part of the variance in the data.
A model might have a negative $R^2$ in case its MSE exceeds the variance of the data, which frequently occurs with models operating on high-dimensional inputs.
Note that comparing MAE and MSE metrics across different output variables can be misleading.
For example, on a variable with very low variance (such as $z$), even a trivial model that returns the variable's average would yield very low MAE and MSE.
Therefore, when comparing across variables, the $R^2$ metric is a better indication of regression performance, which does not depend on the output variable's variance.

Figure~\ref{fig:R2} shows a consistent pattern among all networks: the prediction is best for $x$ and $y$ variables ($R^2 > 0.5$).
$z$, which encodes height, proves harder, and $\theta$, representing the subject's head orientation, results being the most complex variable to estimate.
The estimation of $\theta$ is more sensitive to challenging testing images, such as those where the subject is very far, or looking away from the drone, or only partly visible.
Additionally, the limited image resolution has a higher impact on $z$ and $\theta$ than on $x$ and $y$, as confirmed by the systematic loss in accuracy reducing the input image's size from the $160 \times 16$ to the $80 \times 32$ NN.
Lastly, considering the effect of quantization vs. full precision, it introduces an approximation of the original network, but it also has a beneficial regularization effect, reducing the parameter space.
In the case of the experiment in Figure~\ref{fig:R2}, these effects are well balanced, and the differences between quantized and full precision models are well within noise margins.

\subsection{Onboard performance}\label{sec:onboard_perf}

Our performance investigation starts from Figure~\ref{fig:energy}, where we show the PULP-Frontnet energy efficiency sweeping all the operative points of the GAP8 SoC.
FC and CL frequency are explored with a growing step of \SI{25}{\mega\hertz}, affecting also the minimum voltage required to enable the desired frequencies (VDD growing step \SI{0.05}{\volt}), as shown in Figure~\ref{fig:energy}-A.
The maximum frequencies, and the required VDD, are selected according to the GAP8 SoC datasheet\footnote{https://gwt-website-files.s3.amazonaws.com/gap8\_datasheet.pdf}.
For each network, i.e., PULP-Frontnet 160$\times$32, 160$\times$16, and 80$\times$32, we show the energy efficiency heat-map as the required energy to perform the inference on one frame.
The model 160$\times$32 is the least efficient of the three, as it requires \SI{14.0}{\mega MAC} operations per frame, resulting in \SI{1.28}{\milli\joule /frame} running at its best configuration of FC@25-\SI{50}{\mega\hertz}-CL@\SI{75}{\mega\hertz}.
Both remaining models, i.e., 160$\times$16 and 80$\times$32, show a higher -- and similar -- energy efficiency, due to their reduced number of operations, as much as \SI{4.3}{\mega MAC} and \SI{4.0}{\mega MAC} per frame, respectively.
In Figure~\ref{fig:energy}-B, we highlight the most energy efficient operative points: FC@\SI{25}{\mega\hertz}-CL@\SI{50}{\mega\hertz} for the model 160$\times$16, and FC@50-\SI{75}{\mega\hertz}-CL@\SI{75}{\mega\hertz} for the 80$\times$32 one, consuming \SI{0.58}{\milli\joule /frame} and \SI{0.43}{\milli\joule /frame}, respectively.
The rest of this section will refer to these most energy-efficient configurations to evaluate both power consumption and inference performance. 

\begin{figure}[t]
    \centering
    \includegraphics[width=\linewidth]{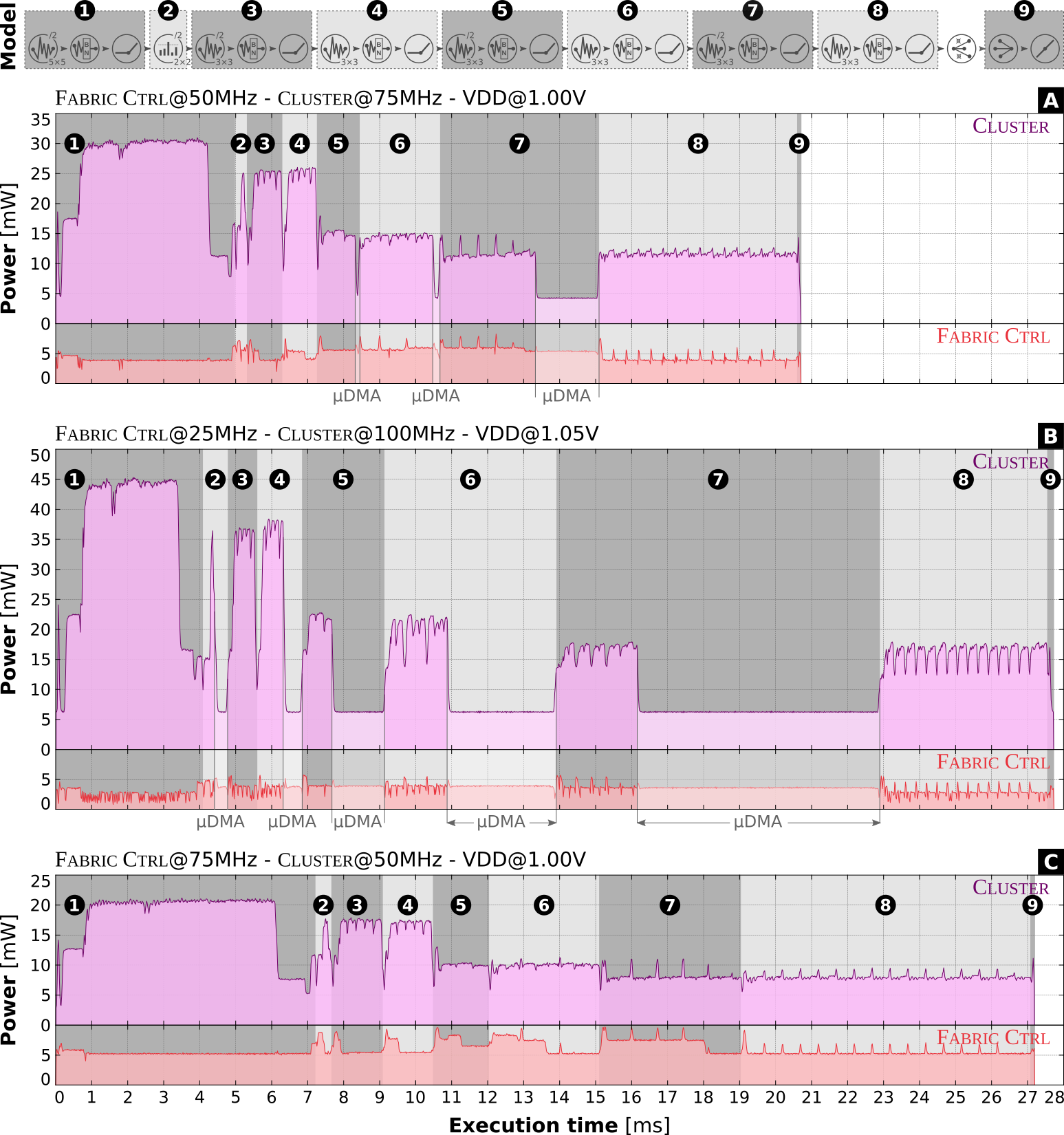}
    \caption{Power traces of \textit{PULP-Frontnet $80\times32$} for one-frame inference in three operative points. Every trace highlights all the computational stages reported in the \textbf{Model}. Measurements are taken after the internal DC/DC converter (i.e., accounting for both \textsc{Fabric Ctrl} and \textsc{Cluster} separately).}
    \label{fig:power_trace}
\end{figure}

Figure~\ref{fig:power_trace} shows the power traces for one-frame inference, reporting both \textsc{Fabric Ctrl} (FC) and \textsc{Cluster} (CL) power domains separately.
Figure~\ref{fig:power_trace}-A refers the to most energy-efficient operative point highlighted in Figure~\ref{fig:energy}-B, i.e., FC@\SI{50}{\mega\hertz}, CL@\SI{75}{\mega\hertz} (80$\times$32).
This configuration exhibits some cluster idleness, up to almost \SI{2}{\milli\second} within layer \circled{7}. 
As reported on the x-axis, the $\mu DMA$ activity (data transfers from the off-chip DRAM to the on-chip L2 memory) does not entirely overlap with the cluster's computation.
As the $\mu DMA$ bandwidth depends on the FC's frequency, it is possible to end up in an operative scenario for which the $\mu DMA$ can not satisfy the data's demand of a faster cluster.
This situation can be even more exacerbated, reducing on one side the FC's frequency and at the same time increasing the CL's one, e.g., FC@\SI{25}{\mega\hertz}, CL@\SI{100}{\mega\hertz} as reported in Figure~\ref{fig:power_trace}-B.
In this case, the CL's idleness increases being well visible for layers \circled{2}, \circled{4}, \circled{5}, \circled{6}, and \circled{7}.
On the contrary, by selecting an operative point such as FC@\SI{75}{\mega\hertz}, CL@\SI{50}{\mega\hertz} (Figure~\ref{fig:power_trace}-C), we can hide all the $\mu DMA$ latencies, obtaining a perfect pipelining between the cluster computation and $\mu DMA$ data transfers, therefore, avoiding any CL's idleness.

Despite the CL's idle time of the first configuration, i.e., FC@\SI{50}{\mega\hertz}, CL@\SI{75}{\mega\hertz}, it results being the most energy-efficient as minimizing the $\mu DMA$ overhead (i.e., the configuration in Figure~\ref{fig:power_trace}-C) brings, as a consequence, a higher mean power consumption of the FC, i.e., \SI{5.9}{\milli\watt} instead of \SI{4.7}{\milli\watt}.
From the energy point of view, the two configurations (Figure~\ref{fig:power_trace}-A vs. C) have a similar energy cost for the CL, i.e., \SI{320}{\milli\joule} and \SI{327}{\milli\joule}, respectively, but a very different one for the FC's domain, as \SI{98}{\milli\joule} vs. \SI{161}{\milli\joule}.
This extra cost ($\sim$\SI{63}{\milli\joule}) is more than $3\times$ the energy overhead for the idleness in the first configuration, turning in a less efficient configuration.
In all power traces, the first computational stage \circled{1} is the most power-hungry due to better utilization of the eight general-purpose cores within the cluster.
Feeding in input the full image allows for an almost-ideal spatial ($width \times height$) parallelization that is not always the case for the remaining convolutional layers, which operate on very small spatial dimensions ($3\times 2$) but very deep input tensor, as in layer \circled{8} with 128 channels, given the PULP-NN spatial parallelization scheme (each core operates on different chunks of spatial data with all input and output channels).

\begin{figure}[t]
    \centering
    \includegraphics[width=\linewidth]{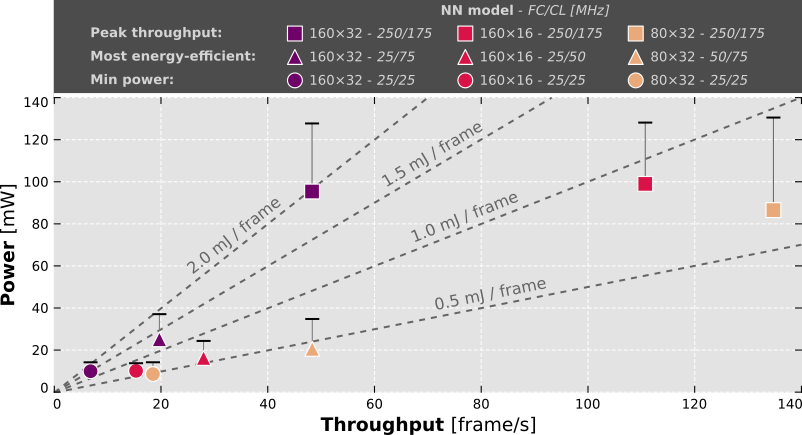}
    \caption{Throughput vs. power consumption for all three NNs, each in three different operative points: \textit{i}) peak throughput, \textit{ii}) the most energy-efficient point highlighted in Figure~\ref{fig:energy}, and \textit{iii}) minimum power consumption. Dashed gray lines show the levels of energy efficiency in \SI{}{\milli\joule}/frame.}
    \label{fig:frame_vs_power}
\end{figure}

Figure~\ref{fig:frame_vs_power} assesses the inference throughput (frame/s) vs. power consumption of the three NN models.
Each model is evaluated in three different configurations: \textit{i}) the one that generates the highest throughput, \textit{ii}) a second one referring to the most energy-efficient operative point identified in Figure~\ref{fig:energy}, and \textit{iii}) a last one for the minimum power consumption.
Each configuration reports both mean and peak power, the latter as a marker on top of each icon.
All three NNs are almost iso-power for a given configuration, showing how the difference in their respective overall energy efficiency comes from a reduced execution time, i.e., increased throughput.
The model PULP-Frontnet 80$\times$32 shows the minimum mean power of \SI{8.6}{\milli\watt} paired with a performance of \SI{18.5}{frame/s}.
The most power-hungry configuration is represented by the model 160$\times$16 running at maximum frequency (FC@\SI{250}{\mega\hertz}, CL@\SI{175}{\mega\hertz}) achieving \SI{110.7}{frame/s} within \SI{99}{\milli\watt}.
On the contrary, the overall peak throughput is given by the model 80$\times$32, reaching \SI{134.7}{frame/s}, with a total SoC power consumption of \SI{86.6}{\milli\watt}.
Lastly, considering the most energy-efficient configuration for all three NNs, the model 80$\times$32 results 2.5$\times$ and 1.7$\times$ faster than the models 160$\times$32 and 160$\times$16, respectively, with very similar power consumption ($\sim$\SI{20}{\milli\watt}).

\begin{figure}[t]
    \centering
    \includegraphics[width=\linewidth]{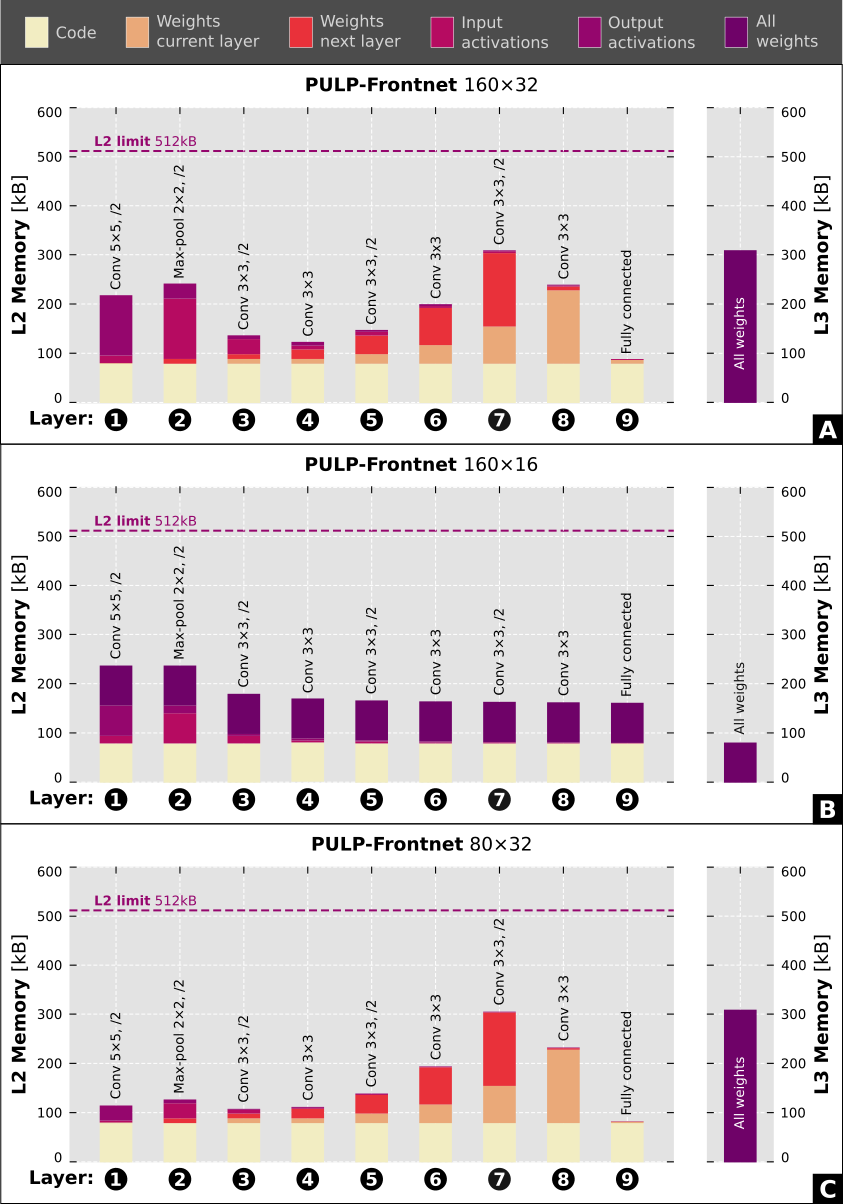}
    \caption{Memory pressure on both L2 and L3 memory for each NN's layer. PULP-Frontnet A) 160$\times$32, B) 160$\times$16, and C) 80$\times$32.}
    \label{fig:memory}
\end{figure}

Figure~\ref{fig:memory} reports the memory pressure on both L2 and L3, for each NN model, namely PULP-Frontnet 160$\times$32, 160$\times$16, and 80$\times$32.
Every sub-figure (A-B-C) shows the amount of L2 memory required at every layer for: \textit{i}) code allocation, \textit{ii}) parameters (i.e., weights) occupancy for the current layer, \textit{iii}) weights allocation for the next layer computation, \textit{iv}) input, and \textit{v}) output buffers (i.e., activations).
The only exception to this representation is given in Figure~\ref{fig:memory}-B where we keep all the NN's weights always in L2 -- due to their small footprint -- avoiding the indication of \textit{current layer} and \textit{next layer} weights allocation.
Each subplot also shows the L2 memory upper-bound of the GAP8 SoC, as a dashed line at \SI{512}{\kilo\byte}.
On the right area of each sub-figure, we also report the total parameters memory footprint located in L3 (i.e., DRAM) that the $\mu DMA$ needs to transfer in L2.
In Figure~\ref{fig:memory}-A/C, during each layer's computation, the $\mu DMA$ transfers the weights required for the next layer, overlapping it with the current computation (i.e., double buffering scheme).
In Figure~\ref{fig:memory}-B all the weights are transferred before the inference starts and are kept available in L2 for the entire application's lifetime, without any additional need to move data from L3.

For all three proposed NNs, the code footprint is almost constant f, i.e., $\sim$\SI{80}{\kilo\byte} and the first layer is the one which requires the largest \textit{output activations} buffer that serves as the input buffer for the second layer.
All three models show a slight pressure on both L2 and L3 memory, as the only potential violation of the L2 upper-bound is represented by the second layer of PULP-Frontnet 160$\times$32, that would require \SI{541}{\kilo\byte} to allocate all weights and intermediate buffers simultaneously in L2.
As already introduced in Section~\ref{sec:background}, the missed opportunity for better exploitation of the L2 memory is a consequence of the deployment tool, i.e., DORY, that is designed to work with NN models characterized by a higher volume of L3 data.
This situation highlights the possibility of the proposed PULP-Frontnet to run together with additional workloads, paving the ground for advanced multi-tasks execution of autonomous navigation algorithms on the GAP8 SoC.

\subsection{In-field control accuracy}\label{sec:in-field}

We conclude the experimental analysis by putting the whole system to the test, i.e., pose estimation task and autonomous navigation, only with the sensory information and computational resources aboard our nano-drone prototype.
We design an experiment where a subject follows a predefined path, while the drone's task is to stay in front of them, at a fixed distance $\Delta=\SI{1.3}{\meter}$.  
We asses the task's quality, using the networks and controller presented in Section~\ref{sec:methodology}.  
We repeat the experiment multiple times using the three proposed NN topologies, and compare how well the drone tracks the target pose in each run.

\begin{figure}[t]
    \centering
    \includegraphics[width=\columnwidth]{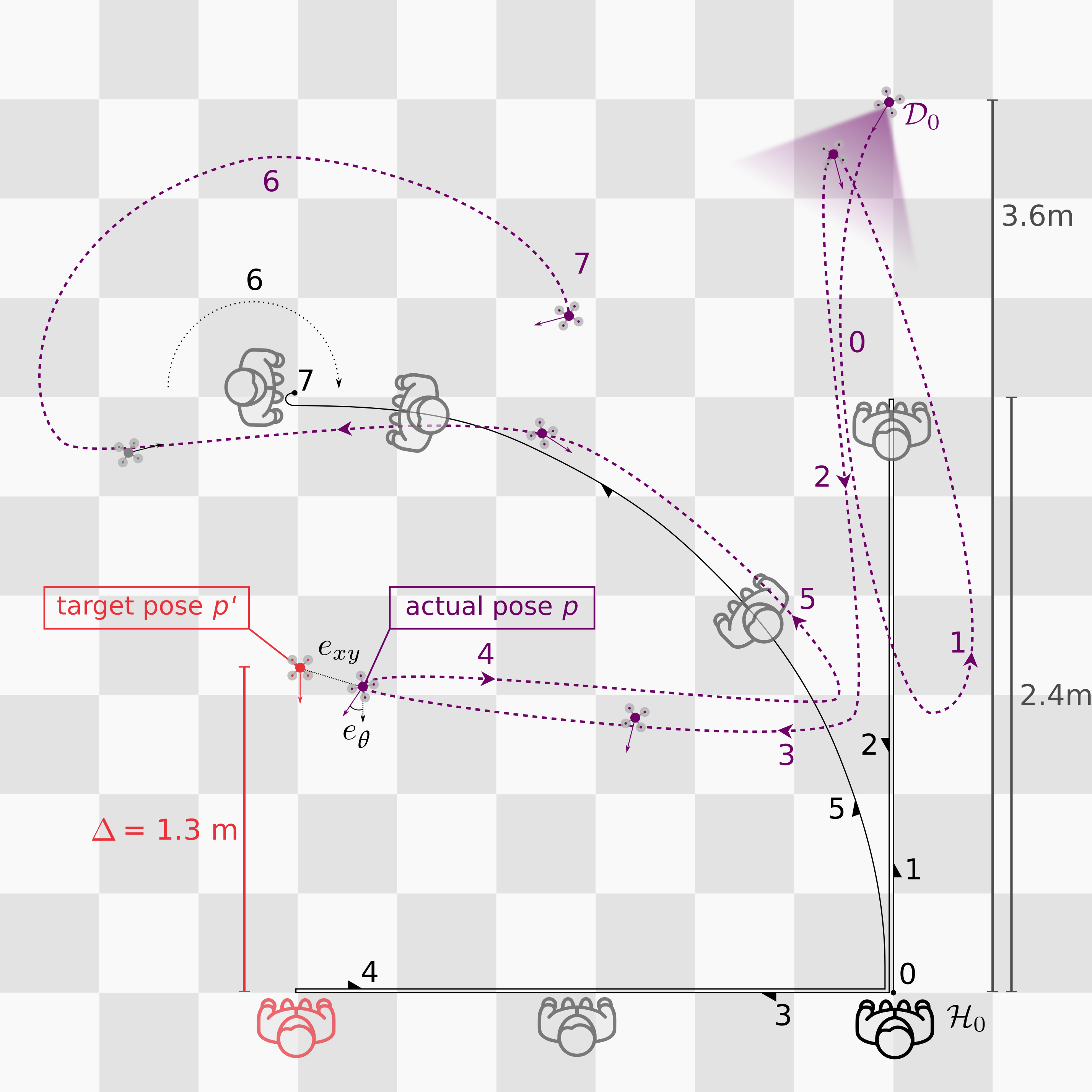}
    \caption{Top-view of the in-field experimental setup (description in text).}
    \label{fig:experimental_setup}
\end{figure}

\textbf{Experimental setup}.
When testing the system, we noticed that subjects tend to adapt their motion to the drone behavior, e.g., a drone reacting slowly and erratically leads them to move slower.
For our experiment, this would make different runs not comparable to each other.
To ensure objective measurements across runs, we ask subjects to completely ignore the drone behavior as they move: this is possible since the drone is very small and subjects don't feel threatened by potential collisions.
To control the subject's motion, we add markers to the floor for each step to be taken: subjects are instructed to step every beat of a metronome, and therefore move at the same speed in every run, independently on the drone behavior.
Figure~\ref{fig:experimental_setup} illustrates the setup and the path that our subjects are instructed to follow. 
The entire pattern takes \SI{50}{\second} among eight phases (0-7), and no pause is taken between them.

\begin{description}[leftmargin=!,labelwidth=10pt]
    \item[Init:] The run starts with the subject standing at pose $\mathcal{H}_0$, facing towards the top side of the map. The drone is initially hovering at pose $\mathcal{D}_0$, \SI{3.6}{\meter} in front of the subject. The drone points \SI{30}{\degree} to the left of the subject; the camera can therefore see the subject, facing towards it. The camera field-of-view is highlighted in Figure~\ref{fig:experimental_setup}.
    \item[Phase 0 (\SI{5}{\second}):] The subject stands still for \SI{5}{\second}; during this time, the drone is expected to rotate left by \SI{30}{\degree} and move forwards to go at \SI{1.3}{\meter} in front of the subject.
    \item[Phases 1/2 (\SI{12}{\second}):] The subject walks forward covering \SI{2.4}{\meter} in \SI{6}{\second}, and then backward for the same time and distance. The drone is expected to track the user by moving backward and then forward.
    \item[Phases 3/4 (\SI{14}{\second}):] Without changing their orientation (i.e., still facing towards the top side of the map), the subject moves sideways towards their left for \SI{2.4}{\meter} and \SI{7}{\second}, and then towards their right.
    \item[Phase 5 (\SI{6}{\second}):] The subject walks along a quarter of a circle, with radius \SI{2.4}{\meter}, counterclockwise, facing the direction of the path, in \SI{6}{\second}. At the end of this phase, the subject faces towards the left of the map.
    \item[Phase 6 (\SI{8}{\second}):] The subject rotates in place, clockwise, by \SI{180}{\degree}, in \SI{8}{\second}; at the end of this phase, the subject faces towards the right of the map. The drone is expected to perform a half circle with \SI{1.3}{\meter} radius while always pointing at its center.
    \item[Phase 7 (\SI{5}{\second}):] The subject stands in place for \SI{5}{\second} (this gives the drone enough time to complete its motion).
\end{description}

Note that the experiment challenges the drone with increasingly difficult tasks: reaching a standing target in phase 0; following a target that moves without changes in orientation in phases 1-4; keeping track of a target that moves and rotates in phase 5; staying in front of a target that spins in place in phase 6.
Note in particular that phases 5-7 test the drone's ability to predict the orientation of the subject's head, which is the most challenging component of the pose to predict (Section~\ref{sec:error_fit}).

For every experimental run, we record the output of inference, and the true poses of both subjects $p^{\mathcal{W}}_{\mathcal{H}}$ and drone $p^{\mathcal{W}}_{\mathcal{D}}$, captured in the world-fixed motion capture frame $\mathcal{W}$.
From these data, in post-processing, we build a dataset consisting of a list of predicted  $\tilde p^{\mathcal{D}}_{\mathcal{H}}$ and ground truth $p^{\mathcal{D}}_{\mathcal{H}}$ poses of the subject relative to the drone.
We run the experiment with two different subjects, neither of whom is part of the training dataset.
Before the experiment, each subject practices the timed movements a few times, as they require some coordination, especially in sudden changes of direction.
For each subject, we perform ten runs: three runs for each of the three networks described in Section~\ref{sec:pulp-frontnet}, in which the model is run at its maximal throughput operative point; and one run, acting as an upper bound of the achievable performance, where the controller uses as input the ground truth relative pose (measured by the motion tracking system), yielding to a total of twenty runs.

\begin{figure}[t]
    \centering
    \includegraphics[width=\columnwidth]{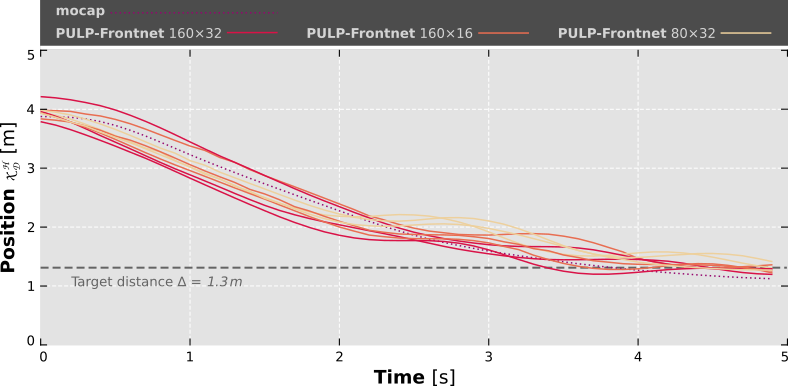}
    \caption{Ground truth $x$ component of the drone pose with respect to the subject during phase 0, for all 10 runs (3 for each network, plus 1 using ground truth relative poses). The dashed gray line represents the target distance $\Delta=\SI{1.3}{\meter}$.}
    \label{fig:trajectory}
\end{figure}

\textbf{Metrics.}
For each run, we measure the following metrics.
To measure the models' ability to interpret the images, we compute the $R^2$ of the prediction with respect to the ground truth for each model's output variable; this metric is comparable with those reported in Section~\ref{sec:error_fit} on the testing set.
In this case, however, the images are acquired during flight, with continuously variable pitch and roll, and with a different distribution of the target variable values, as the drone is actively tracking the subject.
To measure how well the drone is performing the task, we compute statistics on the difference between the drone target pose $p'$ (calculated from the ground truth pose of the subject) and the drone's actual pose $p$ during the entire run.
An example of the two poses at a specific point in time (the end of phase 3) is represented in red and violet, respectively, in Figure~\ref{fig:experimental_setup}.
In particular, we separately quantify:
\begin{itemize}
    \item $e_{xy}$: the horizontal component of the distance between $p$ and $p'$ (absolute position error);
    \item $e_{\theta}$: the difference in orientation between $p$ and $p'$ (absolute angular error).
\end{itemize}
Note that we ignore the $z$ component of the position error, because the target height is approximately constant in our task.

\textbf{Results.}
Figure~\ref{fig:trajectory} illustrates how the drone approaches the subject (who is standing still) during phase 0 of each run.
In this phase, the drone is expected to rotate \SI{30}{\degree} to its left and reach a distance of \SI{1.3}{\meter} from the subject standing at an initial pose approximately \SI{3.6}{\meter} away.
When provided pose estimations from inference, the trajectory converges marginally slower to the target pose than the trajectory observed with perfect estimations.
In particular, minor oscillations in the final part of the trajectory are caused by errors in the distance prediction, depending on the drone pitch.
As the drone gets closer to the target pose, the model detects a decreasing distance, which yields the controller to pitch up to decelerate: the changes in the image due to the different camera pitch lead the model to estimate a slightly higher distance of the subject than previously thought, which pushes the drone to pitch down again to get a little closer; due to the synthetic pitch augmentation technique described in Section~\ref{sec:dataset}, these pitch-dependent errors are small enough that the control is stable; in contrast, experiments on models trained without pitch augmentation show unstable behavior.

\begin{figure}[t]
    \centering
    \includegraphics[width=\linewidth]{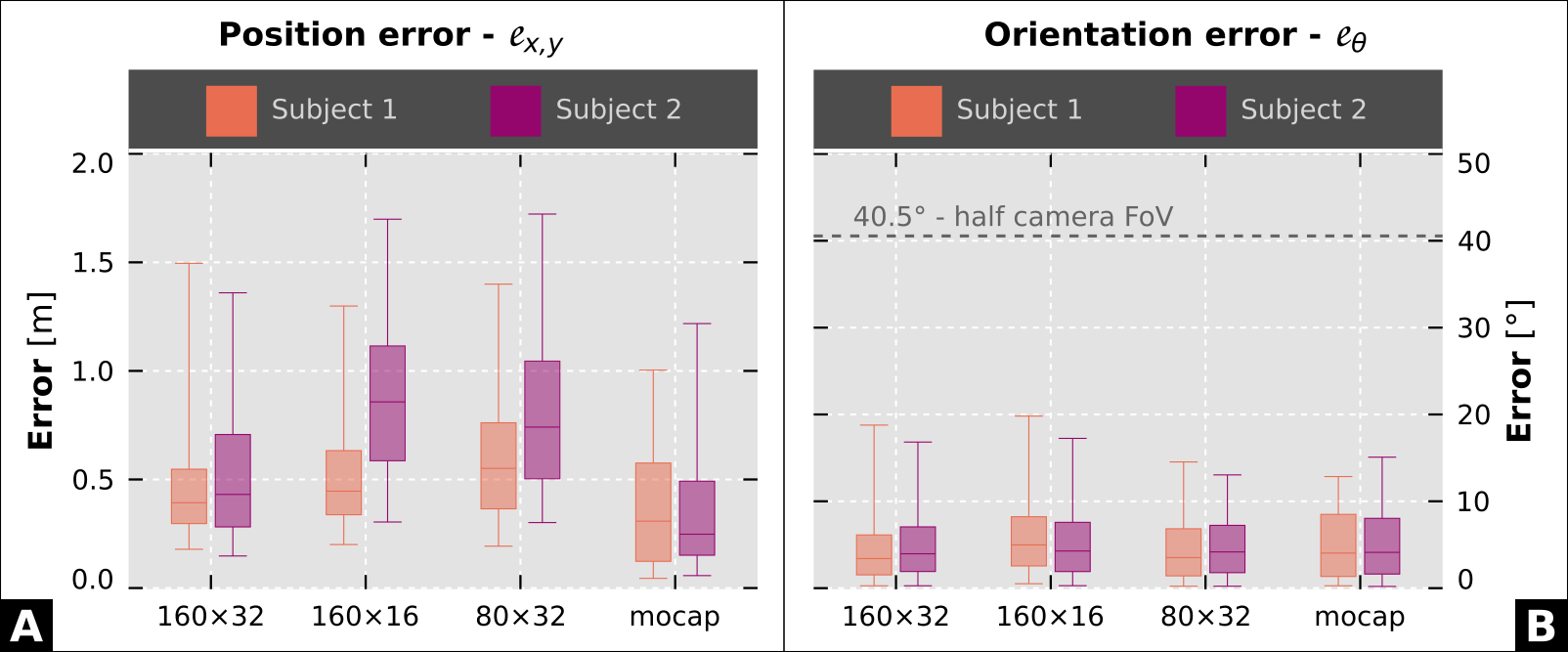}
    \caption{Distribution of control errors for the two subjects (color), using the proposed models. Boxplot whiskers mark $5^{t}h$ and $95^{th}$ percentile of data. The dashed line on the right plot marks the half-field of view of the camera (\SI{40.5}{\degree}): when the angular error is smaller, the subject is visible in the frame.}
    \label{fig:error}
\end{figure}

\begin{table}[h]
\caption{In-field experiment results.}
\begin{center}
\begin{tabularx}{\linewidth}{lXXXXXXXX}
\toprule
\small
\multirow{2}{*}[-0.5\dimexpr \aboverulesep + \belowrulesep + \cmidrulewidth]{Network} & \multirow{2}{*}[-0.5\dimexpr \aboverulesep + \belowrulesep + \cmidrulewidth]{Rate [\SI{}{\hertz}]} & \multicolumn{3}{c}{Regression $R^2$} & \multicolumn{4}{c}{Median pose error} \\
\cmidrule(lr){3-5}\cmidrule(l){6-9}
                & & $x$ & $y$ & $\theta$ & \multicolumn{2}{c}{$e_{xy}$ [\SI{}{\meter}]} & \multicolumn{2}{c}{$e_{\theta}$ [\SI{}{\degree}]}\\
\midrule
$160\times32$   & 48 & 0.87 & \textbf{0.87} & \textbf{0.71} & \multicolumn{2}{c}{\textbf{0.41}} & \multicolumn{2}{c}{\textbf{3.7}}\\
$160\times16$   & 111 & \textbf{0.93} & 0.82 & 0.56 & \multicolumn{2}{c}{0.61} & \multicolumn{2}{c}{4.7}\\
$80\times32$    & \textbf{135} & 0.88 & 0.75 & 0.54 & \multicolumn{2}{c}{0.63} & \multicolumn{2}{c}{4.0}\\
mocap           & 30 & 1.00 & 1.00 & 1.00 & \multicolumn{2}{c}{0.26} & \multicolumn{2}{c}{4.1}\\
\bottomrule
\end{tabularx}
\end{center}
\label{table:in-field}
\end{table}

Table~\ref{table:in-field} reports the rate of the high-level controller on the STM32 (see Section~\ref{sec:control}) and summarizes the metrics over all runs.
The $R^2$ scores indicate good regression performance; in comparison with scores obtained on the testing set (Figure~\ref{fig:R2}), all variables ($\theta$ in particular) are estimated significantly better, for all networks.  For example, network $160\times32$ improves its $R^2$ from $(0.70, 0.67, 0.12)$ to $(0.87, 0.87, 0.71)$ on variables $(x, y, \theta)$.
This improved performance is a consequence of the closed-loop system under test actively following the subject.
In fact, in the in-field tests, the images are acquired mainly from a frontal pose and a close distance, creating a virtuous circle where the better the drone follows the subject, the easier it is to predict the correct pose.
In general, performance trends across various networks and different variables match those observed on the more challenging testing set.

\begin{figure}[t]
    \centering
    \includegraphics[width=\linewidth]{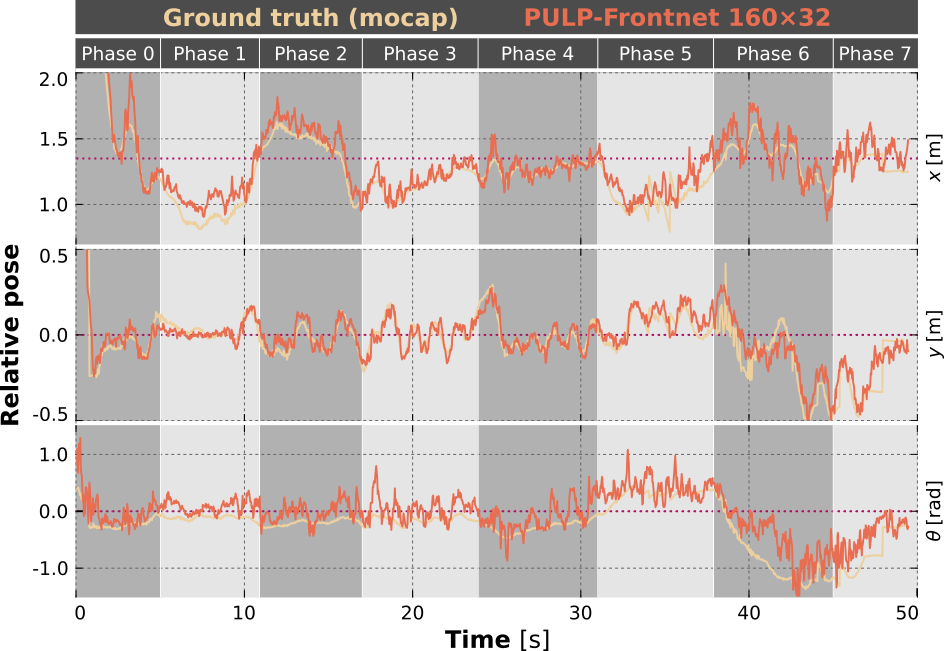}
    \caption{Ground truth (yellow) and network prediction (orange) versus time, during a single experimental run using the 160$\times$32 network, reporting $x$, $y$, and $\theta$ components of the relative pose. The horizontal dashed line (violet) corresponds to the relative pose that the controller attempts to keep. We highlight the time-intervals associated with the different phases of the run.}
    \label{fig:timeseries}
\end{figure}

Table~\ref{table:in-field} also reports the median values for metrics $e_{xy}$ and $e_{\theta}$, and Figure~\ref{fig:error} illustrates their distribution, separately on the two subjects.
For all runs, the angular error $e_{\theta}$ is consistently lower than $\SI{20}{\degree}$, with a median value below $\SI{5}{\degree}$
This shows that, for both subjects, all neural networks effectively estimate the subject's horizontal position to adjust the drone yaw to keep the user in the center of the frame.
In contrast, the position error $e_{xy}$, while generally below \SI{1.0}{\meter}, changes depending on the network and the subject.
The lower bound for this error is given by the mocap runs, in which the controller is fed with the ground truth relative pose of the subject w.r.t. the drone.
We observe that the position error obtained using our best model is less than two times larger than the lower bound.
Lower position errors correspond to networks with higher $R^2$ values, with the 160$\times$32 network performing best on both metrics.
It is important to note that mistakes in estimating the user's head orientation (i.e., on the $\theta$ output variable) are reflected as \emph{position} errors in this experiment.

Figure~\ref{fig:timeseries} compares the network prediction to the ground truth during a single run. 
We observe that, for all pose components, the prediction tracks the ground truth without systematic bias and with good accuracy. 
During the first few seconds (phase 0), $x$ is correctly estimated much above the desired value, which leads the drone to come closer to the user quickly.
For example, during phases 1 and 2,  when the subject walks forward and then backward, the robot moves to keep the distance to the desired value, making the $x$ component reflects the user's movements.
For $\theta$, predictions are noisier than $x$ and $y$, but still manage to capture large-scale patterns in the target variable.
This effect is well visible in the last 10 seconds, i.e.,  phases 6 and 7, where the ground truth of $\theta$ is consistently negative as the drone tracks the user's in-place rotation; the prediction properly also captures this pattern.

\section{Conclusion}\label{sec:conclusion}

In this work, we presented PULP-Frontnet, a novel CNN that visually estimates the pose of a freely-moving human subject, controlling the robot to stay at a constant distance in front of them. 
Solving this HDI problem on an autonomous nano-drone is a challenging and valuable task in the IoT domain.
These robotic helpers can be envisioned as the next-generation ubiquitous IoT devices, ideal for indoor operations near humans.
We propose a general methodology for CNNs' architecture design, dataset collection and augmentation strategies, 8-bit quantization, and deployment on a PULP-based multi-core SoC (i.e., the GWT GAP8).
We consider three CNN variants with different trade-offs on computation, performance, power envelope, and memory needs, running on the GAP8 SoC aboard a COTS Crazyflie 2.1 nano-quadrotor (i.e., 27 grams).
Our results show a remarkable peak performance of \SI{135}{frame/\second} onboard inference rate within only \SI{86}{\milli\watt} power consumption.
Our CNN shows the same regression performance of the resource-unconstrained full-precision baseline, even involving subjects never seen during training.
In-field experiments, made with a fully integrated demonstrator, exhibit excellent control performance (median absolute angular error below \ang{5}) with minimal resource use (down to \SI{4.3}{\mega MAC/frame} operations, and \SI{184}{\kilo\byte} memory footprint).  
With a peak energy efficiency of \SI{0.43}{\milli\joule/frame}, we leave more than enough computational power for additional data analytics aboard our nano-UAVs, paving the way to the ultimate mobile IoT edge-node.

\bibliographystyle{IEEEtran}
\bibliography{IEEEabrv,bibliography}

% that's all folks
\end{document}